\documentclass[]{fairmeta}

\newcommand{\alg}{PCL}

\usepackage{url}            %
\usepackage{booktabs}       %
\usepackage{amsfonts}       %
\usepackage{nicefrac}       %
\usepackage{microtype}      %
\usepackage{xspace}
\usepackage{multirow}
\usepackage{amsmath}
\usepackage{mathtools}
\usepackage{algorithm}
\usepackage{algpseudocode}
\usepackage{color}
\usepackage{MnSymbol}
\usepackage{color, colortbl}
\usepackage{enumitem}
\usepackage{comment}
\usepackage{bm}
\usepackage{wrapfig}
\usepackage{etoc}
\usepackage{makecell}

\title{Prompt Curriculum Learning for \\ Efficient LLM Post-Training}

\author[1,2]{Zhaolin Gao}
\author[1,3]{Joongwon Kim}
\author[2]{Wen Sun}
\author[2]{Thorsten Joachims}
\author[1]{Sid Wang}
\author[1]{\\ Richard Yuanzhe Pang}
\author[1]{Liang Tan}

\affiliation[1]{Meta Superintelligence Labs}
\affiliation[2]{Cornell University}
\affiliation[3]{University of Washington}

\abstract{We introduce \textbf{Prompt Curriculum Learning (\alg{})}, a lightweight reinforcement learning (RL) algorithm that selects intermediate-difficulty prompts using a learned value model to post-train language models. Since post-training LLMs via RL remains sensitive to batching and prompt selection strategies, we first conduct a series of systematic experiments where we (1) determine the optimal training batch size that balances generation efficiency and gradient quality and (2) establish the importance of focusing on prompts of intermediate difficulty for the policy. We build upon these results to design \alg{}, which identifies prompts of intermediate difficulty for the current policy in an on-policy manner by using a value model that is concurrently updated based on the current policy. 
By focusing on informative prompts that yield high effective ratios, \alg{} achieves either the highest performance or requires significantly less time to reach comparable performance to its counterparts. 
Compared to rollout-based filtering methods, \alg{} avoids costly rollouts and achieves $12.1\times$ and $16.9\times$ faster speed on identifying intermediate-difficulty prompts when training on MATH and DeepScaleR, respectively. We further demonstrate that our value model accurately predicts prompt difficulty and allows \alg{} to focus on progressively more challenging prompts during RL. Our results present a new methodology that delivers improved tradeoff between upper-bound performance and efficiency for reasoning-focused RL.}

\date{\today}
\correspondence{Zhaolin Gao at \email{zhaolin@meta.com}}

\begin{document}

\maketitle

\section{Introduction}

Recent large language models (LLMs), such as {OpenAI-o1}~\citep{openaio1} and {DeepSeek-R1}~\citep{deepseekai2025deepseekr1}, have demonstrated strong performance by producing long chain-of-thought (CoT) solutions~\citep{wei2023chainofthoughtpromptingelicitsreasoning, deepseekai2025deepseekr1, zeng2025simplerlzooinvestigatingtamingzero}. A key driver of these improvements is reinforcement learning (RL) with rule-based rewards, using algorithms such as PPO~\citep{schulman2017proximal} and GRPO~\citep{shao2024deepseekmath}. By generating responses online from the current model, RL enables LLMs to self-explore and iteratively improve based on their own outputs.

Substantial effort has been devoted to improving both the performance and efficiency of RL training for LLMs~\citep{brantley2025acceleratingrlllmreasoning, xu2025rolloutsusefuldownsamplingrollouts, Polaris2025, sun2025improvingdataefficiencyllm}. A recurring insight across recent works~\citep{yu2025dapoopensourcellmreinforcement, zhang2025speedrlfastertrainingreasoning, zheng2025actpaysefficientreinforcement} is that training on prompts of intermediate difficulty (i.e., neither too easy nor too hard for the current policy) yields significantly better data efficiency. However, 
existing approaches on identifying intermediate prompts typically rely on either rollouts from the current model or a dictionary that tracks average rewards from previous epochs. The former introduces substantial training overhead due to the high cost of online generation, while the latter suffers from off-policyness especially when the dataset is large. In addition, while these works primarily focus on prompt difficulty, many hyperparameters (e.g., batch size) can significantly affect convergence but remain underexplored in prior work.

In this paper, we conduct a systematic investigation into how prompt selection in conjunction with batch configuration impacts the convergence of RL training. We uncover two key findings. First, \textbf{there exists an optimal batch size that achieves the best trade-off between faster generation time and smaller gradient noise.} While larger batches reduce gradient noise and allow for higher learning rates, they also increase generation time, limiting update frequency. We identify a sweet spot at the transition point between sublinear and linear generation time growth, where convergence speed is maximized. Second, \textbf{prompts of intermediate difficulty are the most effective for learning.} When a prompt is too easy or too hard, gradient signals tend to vanish, leading to wasted compute. In contrast, prompts for which the model has a $\sim$50\% success rate yield the highest gradient norms and require fewer samples to obtain informative updates. We validate this finding empirically across models, datasets, and batch configurations.

Building on these insights, we introduce \textbf{Prompt Curriculum Learning} (\alg), an efficient algorithm that dynamically selects prompts of intermediate difficulty using a value model. At each step, \alg{} samples a large pool of candidate prompts, predicts their expected reward with a single forward pass, and greedily selects those closest to a target threshold (e.g., $0.5$). This approach avoids the overhead of rollout-based prompt filtering while also being much more on-policy than dictionary-based methods. We benchmark \alg{} across a wide range of models and datasets, including Qwen3-Base (1.7B, 4B, 8B) and Llama3.2-it (3B) on MATH, Olympiad-Bench, Minerva MATH, AMC, and AIME. Empirically, \alg{} either achieves the highest performance or requires substantially less training time to reach comparable performance.

\begin{figure}[t]
    \centering
    \includegraphics[trim={0 0 0 0}, clip, width=\textwidth]{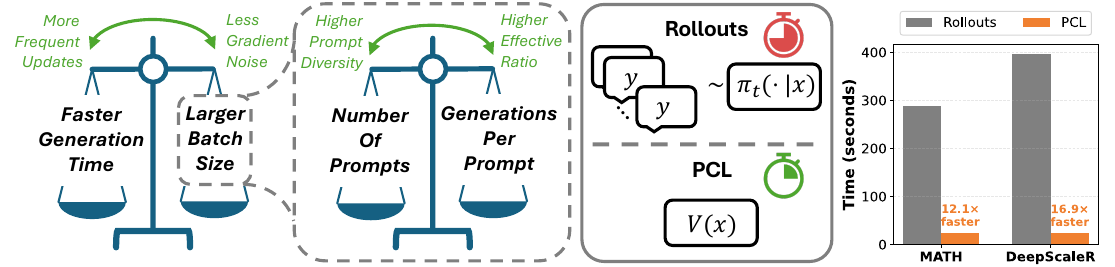}
    \caption{We conduct a systematic investigation of the trade-offs on generation time vs. batch size and number of prompts vs. generations per prompt. We identify an optimal batch size that achieves the best trade-off and discover that the prompts of intermediate difficulty are the most effective for learning. Building on these insights, we introduce \textbf{Prompt Curriculum Learning} (\alg), which trains a value model online for prompt filtering. Compared to the rollout-based filter method, \alg{} is $12.1\times$ and $16.9\times$ faster during prompt filtering when training on MATH and DeepScaleR respectively.}
    \label{fig:main_figure_1}
\end{figure}

\section{Problem Setup}

Let $x$ denote a prompt (e.g., a math question), and let $y$ denote a sampled solution of length $|y|$ generated autoregressively from a policy $\pi$, i.e., $y \sim \pi(\cdot \mid x)$. We assume a binary reward function $r(x, y) \in \{0, 1\}$, where $r(x, y) = 1$ if the final answer in $y$ is correct and $0$ otherwise. Since the reward is binary, we denote $p_\pi(x) \coloneqq \mathbb{E}_{y \sim \pi(\cdot \mid x)}[r(x, y)]$ as the probability of generating a correct answer from policy $\pi$ on prompt $x$, and $A(x, y) \coloneqq r(x, y) - p_\pi(x)$ as the advantage. To optimize $\pi$, we adopt the purely on-policy variant of GRPO~\citep{shao2024deepseekmath, deepseekai2025deepseekr1}, without KL regularization to a fixed reference policy $\pi_{\mathrm{ref}}$~\citep{yu2025dapoopensourcellmreinforcement} and without standard deviation-based advantage regularization~\citep{liu2025understandingr1zeroliketrainingcritical}, by maximizing:
\begin{align}
    \mathbb{E}_{x \sim \mathcal{D},\, y \sim \pi_{t}(\cdot \mid x)} \left[ \frac{1}{|y|} \sum_{l=1}^{|y|} \frac{\pi(y_l \mid x, y_{<l})}{\pi_{t}(y_l \mid x, y_{<l})} A(x, y) \right],
    \label{eq:main_obj}
\end{align}
at time step $t$ for one gradient update where $y_l$ denotes the $l$-th token in the generated sequence $y$, and $\pi_t$ denotes the policy at step $t$. We adopt this formulation to eliminate the off-policyness during updates, clipping heuristics, and additional hyperparameters, which would complicate our analysis in the following section. We note that this is a \textbf{clean} RL objective that has the same gradient as policy gradient and can be directly derived from the original RL objective of maximizing expected reward: $\mathbb{E}_{x \sim \mathcal{D},\, y \sim \pi(\cdot \mid x)}[r(x, y)]$. The derivation is provided in Appendix~\ref{app:problem_setup}.

\section{Preliminary Investigations}
\label{sec:preliminary_investigation}

In this section, we present a set of preliminary experiments that investigate the interplay between convergence, batch size, the number of prompts per batch, and the number of generations per prompt. 
We first define them in detail.

\textbf{Batch size}, denoted by $b$, refers to the total number of prompt–response pairs in a batch. In our purely on-policy setting, this number also corresponds to the total number of pairs used in a single update. The batch size is given by the product of the number of prompts and generations per prompt. Batch size directly affects the \textbf{generation time}, as larger batches require longer to generate.

\textbf{Number of prompts}, denoted by $m$, refers to the number of unique prompts in a batch. This quantity is closely related to the \textbf{prompt diversity}. Increasing the number of prompts improves the diversity of the batch, which in turn reduces gradient noise and stabilizes learning.

\textbf{Generations per prompt}, denoted by $n$, refers to the number of responses generated for each prompt. These responses are used to estimate the expected reward, which is used to compute the advantage. The number of generations per prompt is related to the \textbf{effective ratio}, defined as the proportion of samples in the batch with non-zero advantages, i.e., the proportion of samples that contribute meaningful gradient signals. Increasing $n$ improves the effective ratio. For example, for a particularly challenging prompt, if $n=2$, both responses may be incorrect, leading to zero advantage and zero gradient under the objective in Eq.~\ref{eq:main_obj}. In contrast, for $n=16$ or $32$, it is much more likely that at least one response is correct, resulting in a non-zero advantage and thus useful gradient updates. Therefore, increasing $n$ would result in a more accurate advantage estimation and a higher effective ratio.

\begin{figure}[t]
    \centering
        \includegraphics[trim={0 0 0 0}, clip, width=\textwidth]{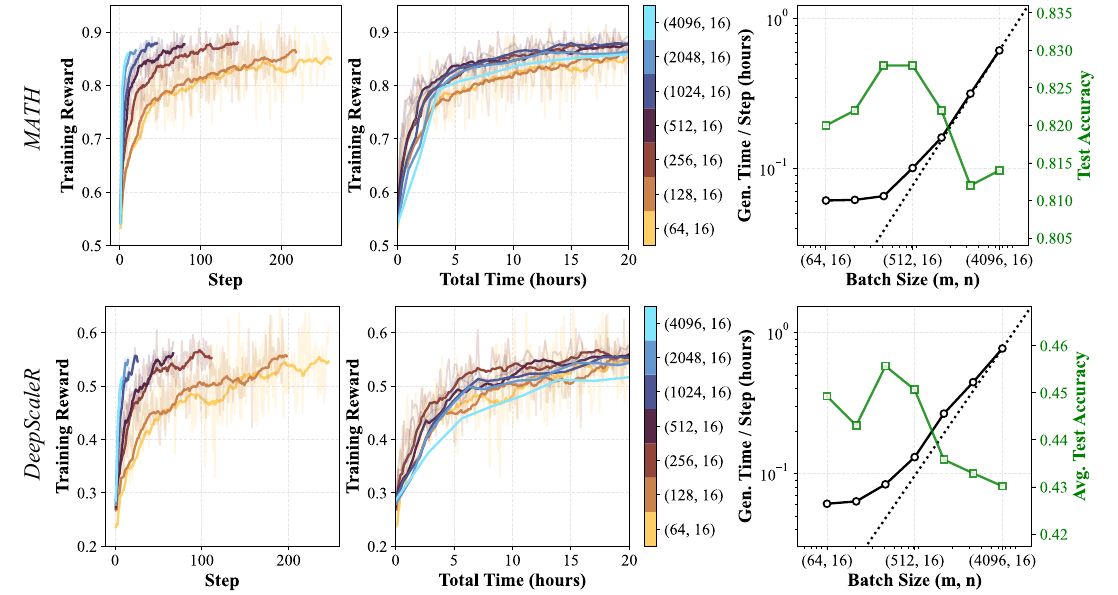}
    \caption{(Left / Middle) Training reward as a function of training steps and wall-clock time for Qwen3-4B-Base on MATH and DeepScaleR. The legend indicates the batch configuration in terms of (number of prompts $m$, generations per prompt $n$). (Right) Generation time per step and test accuracy across different batch sizes. The dashed line represents the linear increase that intercepts the origin and the generation time for the largest batch size. Both axes are in log scale. For key takeaways, refer to the paragraph headers in Section~\ref{sec:optimal_bs}.}
    \label{fig:optimal_bs_1}
\end{figure}

\textbf{Convergence} is defined as the final training or validation reward achieved under a fixed compute and time budget (e.g., number of GPUs and wall-clock time). A method exhibits faster convergence if, under the same computational resources, it reaches a higher reward. Convergence is influenced by generation time, prompt diversity, and effective ratio. Reducing generation time enables more frequent updates, while increasing prompt diversity and effective ratio reduces noise in the gradient and leads to more stable and efficient training.

Overall, these quantities exhibit a natural trade-off. On the one hand, reducing generation time enables more frequent updates within a fixed time budget, allowing the model to train on new rollouts from improved policies. On the other hand, increasing the number of prompts and generations per prompt reduces gradient noise with a higher signal-to-noise ratio. In the following experiments, we perform comprehensive ablations with around $100$K A100 GPU hours to identify the optimal balance between these competing factors.

\subsection{Optimal Batch Size}
\label{sec:optimal_bs}

\begin{figure}[t]
    \centering
    \includegraphics[trim={0 0 0 0}, clip, width=\textwidth]{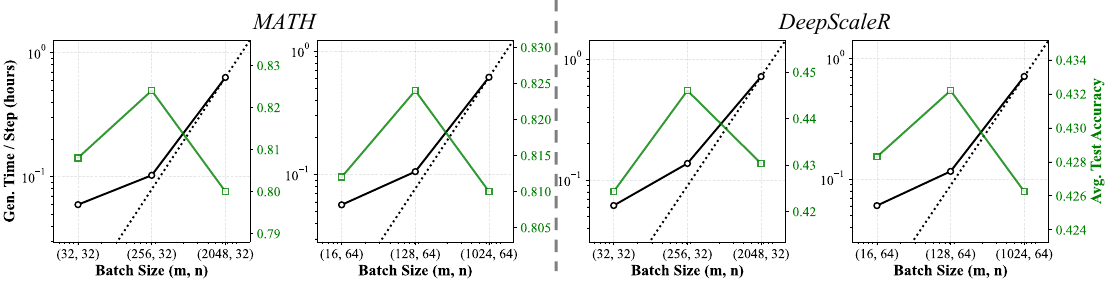}
    \caption{Generation time per step and test accuracy across different batch size combinations (number of prompts $m$, generations per prompt $n$) for Qwen3-4B-Base on MATH and DeepScaleR.}
    \label{fig:optimal_bs_2}
\end{figure}

\textbf{Experiment Setup.} We conduct experiments on both MATH~\citep{hendrycks2021measuringmathematicalproblemsolving} and DeepScaleR~\citep{deepscaler2025} datasets. For MATH, we evaluate on the standard MATH500 split. For DeepScaleR, we include evaluations on MATH500, Minerva Math~\citep{lewkowycz2022solvingquantitativereasoningproblems}, OlympiadBench~\citep{he2024olympiadbenchchallengingbenchmarkpromoting}, as well as competition-level benchmarks including AMC 23, AIME 24, and AIME 25. We report results across four models, Qwen3-1.7B-base, Qwen3-4B-base, Qwen3-8B-base, and Llama3.2-3B-it, covering two model families and a range of sizes. All models are trained with a context length of $4{,}096$ tokens. We use a rule-based reward function based on \texttt{math-verify}~\citep{mathverify2024}, which assigns a reward of +1 for correct ones and 0 for incorrect ones or generations that exceed the context limit. All experiments are implemented using \texttt{Verl}~\citep{Sheng_2025}, a synchronous training setup that alternates between generation and optimization phases. For each batch size, we ablate to find the optimal learning rate with a total of 23 runs. Additional implementation and training details, including learning rate ablations, are provided in Appendix~\ref{app:exp_detail}.

The results for {Qwen3-4B-Base} are presented in Fig.~\ref{fig:optimal_bs_1} and \ref{fig:optimal_bs_2}, including training reward as a function of both training steps and wall-clock time (in hours), generation time per step using \texttt{vLLM}~\citep{kwon2023efficientmemorymanagementlarge}, and test accuracy. For DeepScaleR runs, test accuracy is reported as the average across all six benchmarks. Full results are provided in Appendix~\ref{app:complete_results}.

\textbf{Larger batch sizes converge faster in terms of steps.}  
As shown in Fig.~\ref{fig:optimal_bs_1} (Left), increasing the batch size consistently leads to faster convergence when measured in training steps. This is primarily because larger batches reduce gradient noise, allowing the use of higher learning rates without destabilizing training. The learning rates used in each configuration are listed in Tables~\ref{tab:all_exp_math} and~\ref{tab:all_exp_deepscaler}.

\textbf{Generation time grows sublinearly at first, then linearly.}  
In Fig.~\ref{fig:optimal_bs_1} and~\ref{fig:optimal_bs_2}, we plot generation time per step against batch size, alongside a dashed reference line representing linear growth (intersecting the origin). We observe that generation time initially increases sublinearly with batch size, and transitions to linear growth as batch size continues to increase. This behavior is expected: When the batch size is small, the generation time is dominated by the longest response in the batch. As batch size increases, compute utilization becomes the bottleneck, and generation time scales more linearly.

\textbf{Optimal batch size occurs at the transition point from sublinear to linear scaling.}  
From Fig.~\ref{fig:optimal_bs_1} (Middle / Right) and Fig.~\ref{fig:optimal_bs_2}, there exists a sweet spot in batch size that yields the best convergence speed. Extremely small or large batch sizes lead to suboptimal performance. The optimal point for the fastest convergence tends to lie at the end of the sublinear regime and the beginning of the linear regime in generation time. Specifically, the optimal batch size in our setting is around 8K, achieved with combinations $(m, n) = (512, 16)$, $(256, 32)$, or $(128, 64)$. In other words, the optimal batch size remains fixed, regardless of how it is factorized into $m$ and $n$. We hypothesize that this sweet spot achieves a favorable balance: compared to smaller batch sizes, it can have linearly more generations with sublinear time growth; compared to larger batch sizes, it allows more frequent updates in the same amount of time. To ensure robustness, we validate this phenomenon across different model architectures and sizes, datasets, context lengths, hardware configurations, rollout engines (\texttt{vLLM} vs.~\texttt{SGLang}), and batch configurations. Full results are provided in Appendix~\ref{app:complete_results}. Having established an optimal batch size, the natural question is: \emph{How should we determine the optimal decomposition into the number of prompts and generations per prompt?}

\subsection{Optimal Number of Prompts and Generations per Prompt}
\label{sec:optimal_nm}

\begin{figure}[t]
    \centering
    \includegraphics[trim={0 0 0 5}, clip, width=0.95\textwidth]{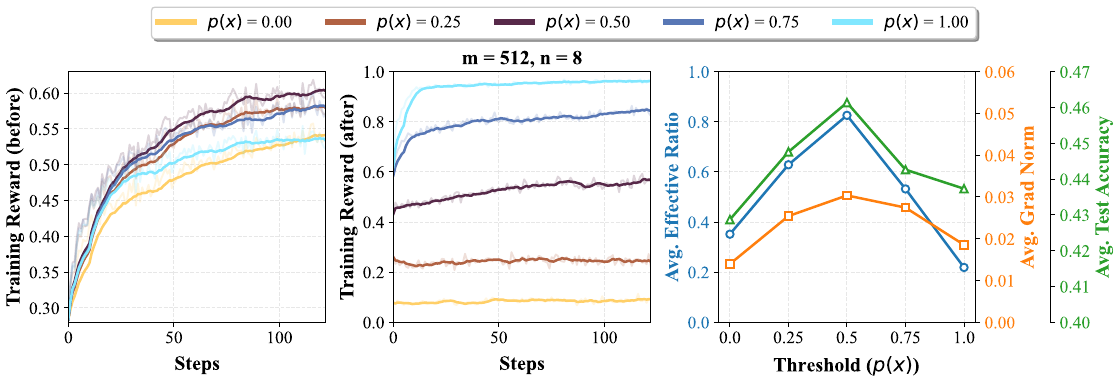}
    \caption{(Left) Training reward before downsampling in terms of step with number of prompts $m=512$ and generations per prompt $n=8$. (Middle) Training reward after downsampling. (Right) Average effective ratio and gradient norm over training steps, and average test accuracy of six benchmarks across different thresholds. For key takeaways, refer to Section~\ref{sec:optimal_nm}.}
    \label{fig:optimal_n_m_2}
    \vskip 0.5cm
\end{figure}

\begin{figure}[t]
    \centering
    \vspace{-0.5cm} 
    \includegraphics[trim={0 0 0 0}, clip, width=0.58\textwidth]{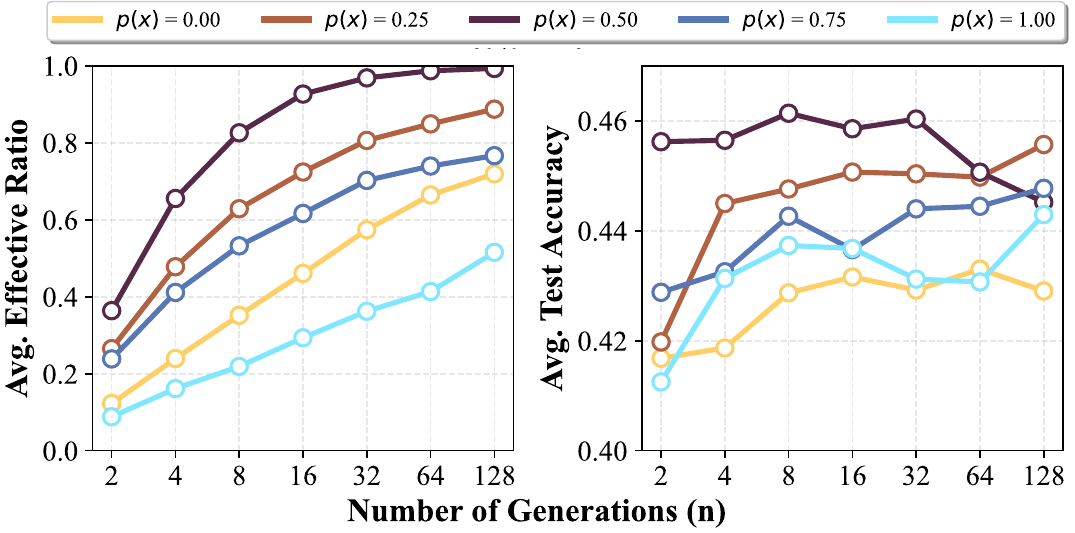}
    \caption{Average effective ratio over training steps and average test accuracy of six benchmarks under different thresholds $p(x)$ and generations per prompt $n$.}
    \label{fig:optimal_n_m_1}
\end{figure}

We hypothesize that the optimal decomposition of the batch size is closely tied to the difficulty of the prompts. Specifically, for extremely easy or difficult prompts, a larger number of generations ($n$) may be necessary to achieve a high effective ratio. In contrast, for prompts of intermediate difficulty ($p(x) \approx 0.5$), fewer generations may be sufficient.

\textbf{Experiment Setup.} 
We use DeepScaleR dataset and Qwen3-4B-Base, and train under different decompositions. To control prompt difficulty, for each batch we first sample $4m$ prompts and generate $4$ responses for each prompt to estimate $p(x)$, similar to \citet{zhang2025speedrlfastertrainingreasoning}. We then perform greedy downsampling to select $m$ prompts that are closest to a specific difficulty threshold $p(x) \in \{0, 0.25, 0.5, 0.75, 1\}$, and sample $n$ generations per selected prompt for training. We are not reusing the $4$ responses to train to avoid selection-induced bias, which keeps the ablation on $n$ comparable. We keep the total batch size fixed at $m \times n = 4096$ and ablate $n$ from 2 to 128. All other experimental configurations remain the same. Full results are shown in Appendix~\ref{app:complete_results}.

\textbf{Downsampling successfully retains target-difficulty prompts.}
As shown in Fig.~\ref{fig:optimal_n_m_2} (Left / Middle), our downsampling procedure effectively retains prompts around the specified threshold. This validates the experimental design and ensures that training focuses on prompts of controlled difficulty.

\textbf{Higher $n$ improves effective ratio and $p(x) = 0.5$ has the highest effective ratio.}
As shown in Fig.~\ref{fig:optimal_n_m_2} (Right) and \ref{fig:optimal_n_m_1}, increasing $n$ consistently improves the effective ratio, and prompts with $p(x) = 0.5$ achieve high effective ratios even with relatively small $n$. For example, the effective ratio for $n = 16$ at $p(x) = 0.5$ is already higher than any other thresholds even with $n = 128$.

\textbf{$p(x) = 0.5$ has the highest gradient norm and test accuracy.}
As shown in Fig.~\ref{fig:optimal_n_m_2} (Right) and \ref{fig:optimal_n_m_1}, training on prompts with $p(x) = 0.5$ yields the highest gradient norms and test accuracy. Interestingly, while increasing $n$ benefits test accuracy for other difficulty levels, we find that for $p(x) = 0.5$, accuracy actually degrades beyond $n = 32$. We suspect this is due to reduced prompt diversity (i.e., smaller $m$), which increases gradient noise despite higher per-prompt sampling. Conversely, based on the previous section, since there exists an optimal batch size, focusing on $p(x) = 0.5$ allows us to use a smaller $n$ and a higher $m$ which improves prompt diversity and also maintains a high effective ratio. In other words, we could have the best of both worlds (effective ratio and prompt diversity) with $p(x)=0.5$. Full results, including ablations across all configurations, are provided in Appendix~\ref{app:complete_results}, and a theoretical connection of the gradient norm and $p(x)$ is provided in Appendix~\ref{app:p_x_grad_norm}.

\section{\alg: Prompt Curriculum Learning}

\begin{algorithm}[t]
\caption{\alg}
\label{alg:pearl}
\begin{algorithmic}[1]
\Require Number of prompts $m$, generations per prompt $n$, threshold $\tau$, sampling parameter $k$
\State Initialize policy $\pi_0$, value network $V^{\pi_{-1}}$
\For{$t=0$ to $T-1$}
    \State Sample a batch with $km$ prompts: $\mathcal{D}_{km}=\{x^i\}_{i=1}^{km} \subset \mathcal{D}$.
    \State Select a batch of $m$ prompts using value model: $\mathcal{D}_m=\underset{S\subseteq \mathcal{D}_{km},\;|S|=m}{\arg\min} \sum_{x \in S} \Bigl|V^{\pi_{t-1}}(x)-\tau\Bigr|.$
    \State Generate for the batch: $\mathcal{D}_m = \bigl\{(x^i,\{y^{i,j}\}_{j=1}^n) \bigr\}_{i=1}^m$ where $y^{i,j} \overset{\mathrm{iid}}{\sim} \pi_t(\cdot\mid x^i)$
    \State Update $\pi_{t}$ to $\pi_{t+1}$ using $\mathcal{D}_m$.
    \State Update $V^{\pi_{t-1}}$ to $V^{\pi_t}$ with loss in Eq.~\ref{eq:value_loss}.
\EndFor
\end{algorithmic}
\end{algorithm}

The previous section demonstrates that prompts of intermediate difficulty ($p(x) \approx 0.5$) are the most sample-efficient for RL training. However, estimating the difficulty of each prompt using actual generations from the policy can be computationally expensive, as the generations for the filtered-out prompts are wasted. To address this issue, we propose a lightweight and efficient alternative: Prompt Curriculum Learning (\textbf{\alg}), which leverages a learned value model during online RL to estimate prompt difficulty using a single forward pass, significantly reducing computational overhead.

At training iteration $t$, we begin by sampling a pool of $km$ candidate prompts from the dataset where $k$ is a hyperparameter. For each prompt $x$, we use a value model to predict its expected reward $V(x)$, which approximates $p_\pi(x) = \mathbb{E}_{y \sim \pi(\cdot \mid x)}[r(x, y)]$. We then greedily select a subset of $m$ prompts whose predicted values are closest to a target difficulty threshold $\tau$ (defaulting to $0.5$), ensuring that the batch is focused on prompts of intermediate difficulty. For each selected prompt, we generate $n$ responses using the current policy and perform standard policy gradient updates. To update the value model, we only use the generated responses and minimize the prediction error between the estimated value $V(x)$ and the empirical average reward across the $n$ generations:
\begin{align}
\label{eq:value_loss}
    \sum_{i=1}^{m} \left( V(x^i) - \frac{1}{n} \sum_{j=1}^{n} r(x^i, y^{i,j}) \right)^2.
\end{align}
This allows us to improve the value model online, without requiring any additional rollouts. Since the value model only takes in the prompt as input which is typically less than 1K tokens in length for math, we find that both training and inference of the value model incur negligible cost and can be completed under $30$ seconds for each step. The full algorithm is summarized in Algorithm~\ref{alg:pearl}. Note that the value model $V$ in our algorithm is one step behind the policy $\pi$, which is acceptable since each update is small with $\pi_{t+1} \approx \pi_t$. We further discuss the alternatives in Section~\ref{sec:discussion}.

\section{Experiments}
\label{sec:experiment}

\textbf{Models \& Datasets.} We use the same sets of models and datasets for experiments as Section~\ref{sec:preliminary_investigation}. We use the same-sized model as the policy for the value model when running \alg. All runs use a 2-day time budget, except for Qwen3-8B-Base on DeepScaleR, which is trained for 3 days. We focus on $m=512$ and $n=16$ as it is one of the best combinations we found in terms of convergence. Unless otherwise noted, \alg{} uses $\tau = 0.5$ and $k = 4$. Similar to~\citet{wang2025reinforcementlearningreasoninglarge} and \citet{zheng2025actpaysefficientreinforcement}, we evaluate the model after training on every $4$K prompts (8 steps), and report the performance of the checkpoint that obtains the best average performance.

\begin{table}[t]\centering
\caption{\textbf{Results on MATH and DeepScaleR.} For each metric, the best-performing method is highlighted in \textbf{bold}, and the second-best is \underline{underlined}. Time is the sum of training and generation time of the checkpoint that achieves the best average performance (excluding validation/checkpointing) in hours. \label{tab:main_result}}
\resizebox{0.76\linewidth}{!}{
\begin{tabular}[t]{c|cc|cc|cc|cc} 
\midrule[0.15ex]
\multirow{2}{*}{\textbf{MATH}} & \multicolumn{2}{c|}{Qwen3-8B-Base} & \multicolumn{2}{c|}{Qwen3-4B-Base} & \multicolumn{2}{c|}{Qwen3-1.7B-Base} & \multicolumn{2}{c}{Llama3.2-3B-it} \\
& {MATH500} & Time & MATH500 & Time & MATH500 & Time & MATH500 & Time \\
\midrule[0.05ex]
$\pi_\mathrm{ref}$ & 72.4 & / & 65.6 & / & 55.4 & / & 42.6 & / \\
GRPO  & 86.4 & 28.3 & \underline{83.0} & 29.2 & 73.6 & 22.0 & 56.2 & \textbf{5.80} \\
Pre-filter & 84.8 & \underline{17.1} & 81.6 & 27.1 & 73.4 & \underline{13.5} & 55.4 & 7.47 \\
DS & \underline{87.8} & 37.8 & 82.6 & 37.1 & \textbf{73.8} & 27.6 & \underline{56.8} & 19.3 \\
SPEED & 81.2 & \textbf{4.25} & 78.8 & \textbf{6.75} & 70.2 & \textbf{1.93} & 42.6 & / \\
GRESO & 87.2 & 29.1 & \underline{83.0} & 33.1 & 73.4 & 17.6 & 56.6 & \underline{7.37} \\
\alg & \textbf{88.2} & 37.2 & \textbf{83.4} & \underline{14.0} & \textbf{73.8} & 24.8 & \textbf{57.8} & 14.3 \\
\midrule[0.15ex]
\end{tabular}
}
\resizebox{0.9\linewidth}{!}{
\begin{tabular}[t]{c|c|cccccc|c|c} 
\midrule[0.15ex]
\multirow{2}{*}{\textbf{DeepScaleR}} & \multirow{2}{*}{Method} & \multirow{2}{*}{MATH500} & \multirow{2}{*}{Olymp.} & Minerva & AMC23 & AIME24 & AIME25 & \multirow{2}{*}{Avg.} & \multirow{2}{*}{Time} \\
 & & & & Avg@4 & Avg@32 & Avg@32 & Avg@32 & & \\
\midrule[0.05ex]
\multirow{6}{*}{{Qwen3-8B-Base}}
& $\pi_\mathrm{ref}$ & 70.2 & 34.3 & 29.8 & 49.1 & 15.8 & 8.8 & 34.7 & / \\
& GRPO & 87.2 & 57.9 & 45.3 & 70.1 & 25.3 & 22.7 & 51.4 & 43.0 \\
& Pre-filter & 86.4 & 54.6 & 44.2 & 69.8 & 26.9 & 22.6 & 50.7 & 67.4 \\
& DS & 87.2 & 55.3 & 45.7 & 71.5 & 24.9 & 24.2 & \underline{51.5} & 69.5 \\
& SPEED & 82.4 & 46.4 & 40.3 & 66.6 & 21.1 & 15.7 & 45.5 & \textbf{19.3} \\
& \alg & 88.4 & 56.2 & 46.8 & 71.2 & 25.2 & 23.9 & \textbf{52.0} & \underline{41.8} \\
\midrule[0.05ex]
\multirow{6}{*}{{Qwen3-4B-Base}}
& $\pi_\mathrm{ref}$ & 65.8 & 34.4 & 26.9 & 47.3 & 10.9 & 7.1 & 32.1 & / \\
& GRPO & 83.4 & 51.0 & 40.1 & 60.7 & 16.1 & 20.7 & 45.3 & 45.5 \\
& Pre-filter & 83.4 & 47.8 & 40.0 & 60.2 & 18.8 & 16.2 & 44.4 & 39.0 \\
& DS & 83.2 & 51.6 & 41.2 & 62.4 & 18.5 & 18.0 & \textbf{45.8} & 40.1 \\
& SPEED & 79.4 & 45.4 & 38.3 & 60.3 & 15.7 & 14.5 & 42.3 & \textbf{10.7} \\
& \alg & 83.0 & 50.6 & 40.9 & 60.8 & 19.4 & 19.4 & \underline{45.7} & \underline{32.8} \\
\midrule[0.05ex]
\multirow{6}{*}{{Qwen3-1.7B-Base}}
& $\pi_\mathrm{ref}$ & 57.0 & 23.9 & 21.8 & 29.0 & 3.8 & 1.1 & 22.8 & / \\
& GRPO & 72.4 & 37.7 & 31.2 & 44.9 & 11.2 & 6.7 & 34.0 & 46.2 \\
& Pre-filter & 74.0 & 36.5 & 32.6 & 45.6 & 11.7 & 7.8 & \underline{34.7} & 44.2 \\
& DS & 73.2 & 36.9 & 31.9 & 42.7 & 10.8 & 7.7 & 33.9 & 41.7 \\
& SPEED & 73.0 & 34.4 & 30.2 & 37.2 & 9.2 & 7.1 & 31.8 & \textbf{22.7} \\
& \alg & 74.4 & 35.6 & 31.5 & 46.3 & 12.5 & 9.2 & \textbf{34.9} & \underline{23.3} \\
\midrule[0.05ex]
\multirow{6}{*}{{Llama3.2-3B-it}}
& $\pi_\mathrm{ref}$ & 42.8 & 12.3 & 13.8 & 19.7 & 4.6 & 0.4 & 15.6 & / \\
& GRPO & 55.2 & 23.1 & 22.6 & 40.0 & 13.3 & 0.0 & 25.7 & 47.5 \\
& Pre-filter & 56.8 & 24.5 & 23.3 & 35.5 & 16.5 & 0.7 & 26.2 & 44.8 \\
& DS & 57.2 & 23.3 & 24.1 & 37.1 & 17.5 & 1.0 & \textbf{26.7} & 40.6 \\
& SPEED & 51.4 & 20.2 & 20.1 & 32.0 & 10.6 & 0.8 & 22.5 & \textbf{3.86} \\
& \alg & 58.8 & 23.9 & 24.0 & 35.2 & 15.0 & 2.1 & \underline{26.5} & \underline{28.7} \\
\midrule[0.15ex]
\end{tabular}
}
\vskip -0.3cm
\end{table}

\textbf{Baselines.} We compare \alg{} against five baselines. We include original \textbf{GRPO}, which performs no prompt filtering and uniformly samples prompts from the dataset. This serves as a standard baseline to assess the impact of filtering strategies. \textbf{Pre-filter} is a heuristic approach that leverages a fixed reference policy $\pi_{\mathrm{ref}}$ to estimate prompt difficulty and filters out easy or hard prompts. \textbf{Dynamic-sampling (DS)}~\citep{yu2025dapoopensourcellmreinforcement} uses $n$ rollouts per prompt to estimate $p_\pi$ for $km$ prompts and filters out prompts with $\hat{p}_\pi=0 \ \mathrm{or}\ 1$. \textbf{SPEED}~\citep{zhang2025speedrlfastertrainingreasoning} improves upon DS by first using $n_\mathrm{init}$ rollouts to estimate where $n \geq n_\mathrm{init}$. It then performs filtering and generates the remaining $n - n_\mathrm{init}$ rollouts. \textbf{GRESO}~\citep{zheng2025actpaysefficientreinforcement} keeps a dictionary of historical rewards based on generations from previous epochs and skips uninformative prompts using the dictionary. We tested GRESO on MATH but not on DeepScaleR, as DeepScaleR is large and limits the training to around 1 epoch under the compute budget which prevents the use of dictionary-based methods. DS, SPEED, and GRESO all keep sampling and generating until there is a full batch. Additional experiment details, including pseudo-codes and hyperparameters, are in Appendix~\ref{app:main_exp_detail}.

\subsection{Convergence Comparison}

\textbf{\alg{} either achieves the highest performance or requires significantly less training time to reach comparable performance.}  
The main results are summarized in Tables~\ref{tab:main_result}. Compared to prior baselines, \alg{} consistently achieves the highest performance across all four models trained on the MATH dataset, and ranks either the first or second using the average of six benchmarks when trained on DeepScaleR. DS requires significantly more time to converge, as it performs generation for all $km$ prompts at each step with $n$ generations per prompt. SPEED's efficient implementation pre-generates $n_{\mathrm{init}}$ rollouts at an earlier step with an old policy and uses them at the current step, treating them as if sampled from the current policy. While this approach reduces generation cost for estimating $\hat{p}_\pi$, it introduces severe off-policyness as the current policy is unlikely to generate those rollouts. We observe that most of the SPEED runs crashed within a few hours, leading to lower convergence time as it would crash afterward. On the other hand, GRESO also suffers from a high degree of off-policyness where the historical estimates are based on outdated policies from the last epoch and may not reflect the current model's performance, especially when the dataset is large. An ablation on the size of the value model is included in Appendix~\ref{app:value_model_size}.

\begin{figure}[t]
    \centering
    \includegraphics[trim={0 0 0 0}, clip, width=\textwidth]{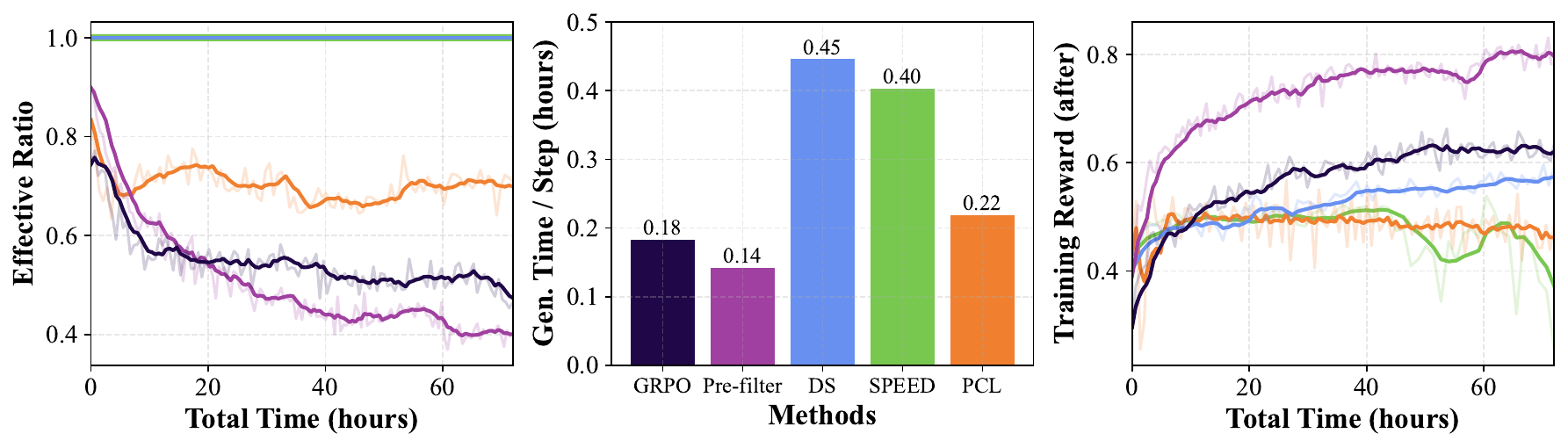}
    \caption{Experiment on DeepScaleR with Qwen3-8B-Base. (Left) Effective ratio w.r.t. training time across five methods. Refer to the middle plot for legend. (Middle) Average generation time per step throughout the training. (Right) Training reward after downsampling. \alg{} either has a higher effective ratio or a lower generation time, and is consistently training on $p(x)=0.5$ prompts.}
    \label{fig:main_results_1}
\end{figure}

\subsection{Analysis \& Ablation}

\textbf{\alg{} consistently achieves either a higher effective ratio or a lower generation time, while maintaining a focus on $p(x) = 0.5$ prompts.} To better understand the training dynamics of each method, we visualize the effective ratio, generation time per step, and training reward after filtering in Fig.~\ref{fig:main_results_1} when training Qwen3-8B-Base on DeepScaleR. \alg{} consistently maintains a higher effective ratio compared to {GRPO} and {Pre-filter}. While {DS} and {SPEED} achieve an effective ratio of $1$ due to resampling, they require significantly higher generation time, with relative increase of 105\% and 81.8\% for DS and SPEED respectively. The slightly higher generation time of \alg{} compared to {GRPO} and {Pre-filter} is that harder prompts require longer generations, and, when the average accuracy of the model on the training set is higher than $0.5$, \alg{} focuses on harder prompts than those two methods. Interestingly, the effective ratio for Pre-filter starts higher than GRPO but quickly drops below. This behavior comes from how {Pre-filter} selects prompts: it excludes very difficult ones based on $\pi_\mathrm{ref}$. As the policy improves during training, many previously difficult prompts transition into the intermediate-difficulty range (e.g., $p(x) \approx 0.5$) for the current model. However, because these prompts were previously filtered out, they are never revisited, causing {Pre-filter} to keep training on easy prompts from the perspective of the current policy. In addition, as shown in Fig.~\ref{fig:main_results_1} (Right), \alg{} consistently focuses on intermediate-difficulty prompts throughout training (the training reward of \alg{} after filtering stays closely to 0.5), whereas other methods gradually shift toward easier prompts as the policy improves which is suboptimal based on the findings in Section~\ref{sec:preliminary_investigation}.

\begin{figure}[t]
    \centering
    \begin{minipage}[t]{0.64\textwidth}
        \centering
        \includegraphics[width=\linewidth, trim={0 0 0 0}, clip]{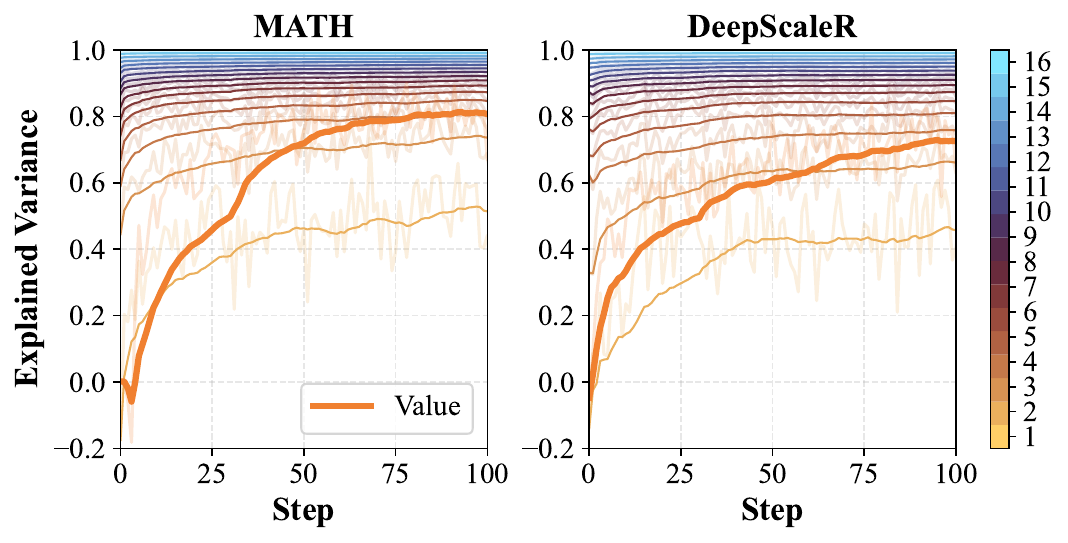}
        \caption{Explained variance of \alg{}'s value model using 16 generations as the ground-truth difficulty ($p(x)$), and the explained variance using 1 to 16 generations to predict the difficulty ($\hat{p}(x)$) on MATH and DeepScaleR with two Qwen3-1.7B-Base models as policy and value model. The accuracy of the value model is similar to using around 3 generations to estimate.}
        \label{fig:main_results_2}
    \end{minipage}\hfill
    \begin{minipage}[t]{0.32\textwidth}
        \centering
        \includegraphics[width=\linewidth, trim={0 0 0 0}, clip]{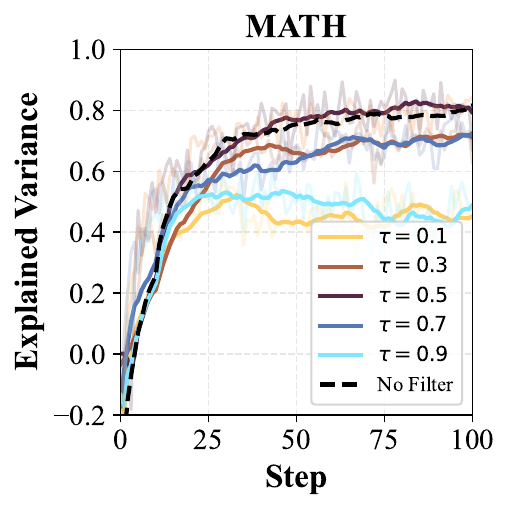}
        \caption{Explained variance on MATH with {Qwen3-1.7B-Base} for \alg{}'s value model with filtering on different thresholds ($\tau$) and without filtering.}
        \label{fig:main_results_3}
    \end{minipage}
\end{figure}



\textbf{The accuracy of the value model is similar to using 3 generations to estimate.} To investigate the prediction accuracy of the value model, we compute the explained variance using the average reward of 16 generations as the ground-truth difficulty $p(x)$. The explained variance is calculated as:
\begin{align}
    1 - \frac{\operatorname{Var}\big(\{p(x^i)-V(x^i)\}_{i=1}^m\big)
}{\operatorname{Var}\big(\{p(x^i)\}_{i=1}^m\big)}
\end{align}
where $\operatorname{Var}$ denotes the variance. In addition, we also use the average reward of 1 to 16 generations as the predicted difficulty $\hat{p}(x)$ and compute their explained variance. The explained variances are computed on prompts randomly sampled from the dataset before filtering. The results on MATH and DeepScaleR with two Qwen3-1.7B-Base models as policy and value models are shown in Fig.~\ref{fig:main_results_2}. Since the prediction head of the value model is randomly initialized, the initial explained variance is very low. As training progresses, the value model improves steadily and achieves an explained variance comparable to using three rollouts per prompt for value estimation. Specifically, with $km = 2048$, generating $n = 3$ rollouts per prompt takes 288 seconds on MATH and 396 seconds on DeepScaleR per step. In contrast, training and inference with the value model require only 23.9 and 23.5 seconds respectively, achieving a $12.1\times$ speedup on MATH and a $16.9\times$ speedup on DeepScaleR. Results are visualized in Figure~\ref{fig:main_figure_1}.

\textbf{The accuracy of the value model with filtering at $\tau = 0.5$ matches that of training without filtering.} One might expect the value model to suffer from filtering, as the training data is biased toward prompts with estimated difficulty near the threshold $\tau$, potentially limiting generalization. To investigate how the choice of threshold $\tau$ affects the accuracy of the value model, we ablate over $\tau \in \{0.1, 0.3, 0.5, 0.7, 0.9\}$. In addition, we train a baseline value model without any prompt filtering (i.e., GRPO but with a value model trained alongside the policy) using {Qwen3-1.7B-Base} for both the policy and value model on the MATH dataset. Results are presented in Fig.~\ref{fig:main_results_3}. We observe that the value model achieves the highest prediction accuracy when $\tau = 0.5$, with performance degrading as the threshold deviates further from 0.5 in either direction. Notably, the accuracy of the value model at $\tau = 0.5$ is comparable to the no-filtering baseline, despite training on a filtered subset of prompts. We hypothesize that filtering at $\tau = 0.5$ still captures a diverse set of reward outcomes, as it is the midpoint of the binary rewards. Moreover, if the average reward of the policy over the training data is not 0.5 (i.e., there is label imbalance), filtering around $\tau = 0.5$ may implicitly rebalance the data, thus improving generalization. In contrast, filtering with extreme $\tau$ values (e.g., $\tau = 0.1$ or $\tau = 0.9$) selects only very easy or very hard prompts, leading to severe label imbalance and reduced predictive accuracy. A deeper theoretical understanding of why $\tau = 0.5$ leads to such effective value model training is an interesting direction for future work.

\begin{wrapfigure}{r}{0.35\textwidth}
    \centering
    \vspace{-0.5cm} 
    \includegraphics[trim={0 0 0 0}, clip, width=0.35\textwidth]{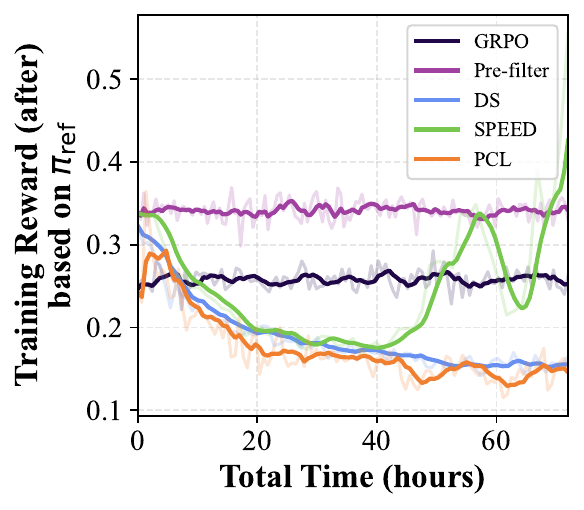}
    \caption{Training reward of \alg{} after filtering based on $\pi_{\mathrm{ref}}$ w.r.t. training time with DeepScaleR and Qwen3-8B-Base. \alg{} progressively focuses on harder prompts during training, despite a fixed threshold of $\tau = 0.5$.}
    \label{fig:main_results_4}
\end{wrapfigure}

\textbf{\alg{} progressively focuses on harder prompts during training, despite a fixed threshold of $\tau = 0.5$.}
To better understand the training dynamics of \alg{}, we analyze how the difficulty of selected prompts evolves over time. Specifically, we use the initial reference policy $\pi_{\mathrm{ref}}$ to generate 16 responses for each prompt in DeepScaleR and compute the average reward, which serves as a proxy for prompt difficulty (i.e., lower average rewards indicate harder prompts). During training on {Qwen3-8B-Base} with \alg{}, we log the average $\pi_{\mathrm{ref}}$-based reward for the filtered prompts at each training step. The results are shown in Fig.~\ref{fig:main_results_4}. For methods that do not perform prompt filtering ({GRPO} and {Pre-filter}), this average remains nearly constant, as these methods uniformly sample from the dataset. In contrast, for methods that apply filtering ({DS}, {SPEED}, and \alg{}), we observe a consistent downward trend in the $\pi_{\mathrm{ref}}$-based reward of selected prompts. This indicates that these methods focus on increasingly harder prompts as training progresses. Although \alg{} maintains a fixed difficulty threshold of $\tau = 0.5$, as the policy improves, previously hard prompts would now appear intermediate (i.e., $\tau \approx 0.5$), allowing \alg{} to continually shift toward more challenging examples. 


\section{Related Work}
\textbf{LLM Post-training.} Reinforcement learning (RL) has become a standard for post-training LLMs, including Reinforcement Learning from Human Feedback (RLHF) \citep{ouyang2022traininglanguagemodelsfollow, christiano2023deepreinforcementlearninghuman, openai2024gpt4technicalreport, kimiteam2025kimik15scalingreinforcement}, enabling the LLMs to generate faithful and harmless responses that closely follow the instruction, and Reinforcement Learning with Verifiable Rewards (RLVR)~\citep{openaio1, yang2024qwen25mathtechnicalreportmathematical, deepseekai2025deepseekr1, yang2025qwen3technicalreport, lambert2025tulu3pushingfrontiers, kimiteam2025kimik2openagentic}, improving model reasoning capabilities using verifiable rewards. These methods typically use algorithms include PPO~\citep{schulman2017proximal}, GRPO~\citep{shao2024deepseekmath}, DR-GRPO~\citep{liu2025understandingr1zeroliketrainingcritical}, OREO~\citep{wang2024offline}, DQO~\citep{ji2024enhancing}, and VinePPO~\citep{kazemnejad2025vinepporefiningcreditassignment}.

\textbf{Efficient RL for LLM Post-training.} Given the huge parameter size for LLMs, there is a large body of work recently focusing on developing more efficient algorithms and data selection methods to enable more efficient RL training for LLMs. Algorithmically, DPO~\citep{rafailov2024directpreferenceoptimizationlanguage}, RAFT~\citep{dong2023raftrewardrankedfinetuning}, REBEL~\citep{gao2024rebel}, REFUEL~\citep{gao2025regressingrelativefutureefficient}, $A^\star$-PO~\citep{brantley2025acceleratingrlllmreasoning}, RAFT++~\citep{xiong2025minimalistapproachllmreasoning}, RLOO~\citep{ahmadian2024basicsrevisitingreinforcestyle}, and REINFORCE++~\citep{hu2025reinforceefficientrlhfalgorithm} are all trying to construct new objective functions that either reduces the number of models used (e.g. value model, reference model, reward model) or reduces the number of generations required for online RL. Another line of works~\citep{xia2024lessselectinginfluentialdata, muennighoff2025s1simpletesttimescaling, ye2025limoreasoning, muldrew2024activepreferencelearninglarge, das2025activepreferenceoptimizationsample, wang2025reinforcementlearningreasoninglarge, sun2025improvingdataefficiencyllm, wang2025infinitesamplingefficientstable, lin2025cppoacceleratingtraininggroup} focuses on improving data selections by reducing the amount of training data to be more sample efficient. DAPO~\citep{yu2025dapoopensourcellmreinforcement} and VAPO~\citep{yue2025vapoefficientreliablereinforcement} resample and keep generating until the effective ratio of the batch is 1 during each step of RL training. However, the generations for a prompt that are either all correct or incorrect are wasted. SPEED~\citep{zhang2025speedrlfastertrainingreasoning} improves on top of these methods by using a smaller number of generations to estimate the effective ratio and only generate the rest of the generations if the existing ones are not all correct or incorrect. GRESO~\citep{zheng2025actpaysefficientreinforcement}, on the other hand, avoids rollouts by using a dictionary-based approach and recording the historical average reward for each prompt from the last epoch. However, it would suffer from off-policyness especially when the dataset is large. 

Our method is the combination of the best of both worlds where \alg{} directly avoids costly rollouts and also is on-policy. Our method is closely related to a classic class of machine learning techniques, Curriculum Learning~\citep{10.1145/1553374.1553380}. Previous works have explored curriculum learning for LLM post-training~\citep{lee2024instructiontuninghumancurriculum, wen2025lightr1curriculumsftdpo, shi2025efficientreinforcementfinetuningadaptive} by either training on progressively harder prompts ordered before training or focusing on certain difficulty range on the fly during RL. Our work falls in this group by always focusing on intermediate difficulty prompts for the current policy.

\section{Discussions \& Conclusion}
\label{sec:discussion}

\alg{} accelerates RL post-training by targeting two findings from our study: (1) there exists an optimal total batch size at the transition between sublinear and linear generation-time scaling, and (2) prompts of intermediate difficulty ($p(x)\approx0.5$) yield the highest gradient signal and sample efficiency. It trains a value model online to identify such prompts, avoiding the wasted rollouts of generation-based filtering (DS, SPEED) and the off-policyness of dictionary-based methods (GRESO). \alg{} either achieves the highest performance or requires significantly less training time to reach comparable performance.

While our experiments focus on binary correctness rewards, \alg{} naturally extends to non-binary scalar rewards. Since the value model $V(x)$ estimates $\mathbb{E}_{y\sim\pi(\cdot|x)}[r(x,y)]$, non-binary $r(x,y)$ only changes its range and the meaning of the target threshold $\tau$. In addition, we note that \alg{} alternates updates between the policy and the value model, meaning that $V^{\pi_{t}}$ is always one step behind the current policy $\pi_{t+1}$. In practice, this lag does not hinder performance, as the per-step policy updates are small with $\pi_t \approx \pi_{t+1}$. We also experimented with using importance sampling to correct for this lag by reweighting based on $\pi_{t+1}(y \mid x) / \pi_{t}(y \mid x)$, but it does not improve the accuracy of the value model and computing $\pi_{t+1}(y \mid x)$ is computationally expensive as $y$ is thousands of tokens long.

We also highlight that prompt filtering methods rely on an implicit assumption of prompt-level generalization: training on a selected subset of prompts will improve performance on the filtered-out ones. For example, \alg{} assumes that training on intermediate-difficulty prompts leads to improvements on both easier and harder prompts, while {DS} and {SPEED} assume that gradually solving not-too-hard prompts enables the model to eventually handle harder ones. While this assumption holds in domains like math where problems often share structural similarities, it may not generalize to other domains. As such, filtering may bias the training distribution and hinder generalization to prompts outside the selected subset.

\section{Limitations}

While \alg{} demonstrates strong empirical performance across a range of models and datasets, our study has several limitations that open avenues for future work.

\textbf{Purely on-policy setting.}  
Our experiments are conducted entirely in a purely on-policy RL setting, where new generations are sampled after each policy update. While this simplifies the analysis and avoids additional hyperparameters (e.g., clipping), it may reduce the generalization to more complex training pipelines that leverage off-policy data or replay buffers.

\textbf{Focus on synchronous setting.}  
Our preliminary investigation and \alg{} are evaluated in a synchronous training setup where data generation and policy updates are alternated step-by-step. However, many large-scale RL pipelines for LLMs adopt asynchronous architectures for better throughput~\citep{wu2025llamarldistributedasynchronousreinforcement,fu2025areallargescaleasynchronousreinforcement}. Extending our analysis and \alg{} to asynchronous settings may require more sophisticated value model training and prompt selection strategies to handle stale or partially updated policies.

\textbf{Relatively short context lengths.}  
We limit our experiments to a maximum context length of 4{,}096 tokens due to compute constraints. While this setting is sufficient for the datasets used (e.g., MATH, DeepScaleR), real-world LLM deployments often involve much longer contexts. From our analysis, for longer context length, the batch size that transitions from sub-linear to linear generation time is larger. Future work could explore the interplay between prompt difficulty, batch decomposition, and context length in long-context regimes.

\textbf{Limited training horizon.}  
Our experiments are constrained to relatively short training runs (e.g., 2–3 days), which may not fully capture long-term convergence behavior, especially for larger models and datasets. Although we observe strong early-stage performance, it remains an open question whether our analysis in Section~\ref{sec:preliminary_investigation} would generalize to much longer training runs.

\clearpage
\newpage
\bibliographystyle{assets/plainnat}
\bibliography{paper}

\clearpage
\newpage
\beginappendix

\addcontentsline{toc}{part}{Appendix}
\etocsetnexttocdepth{3}
\localtableofcontents
\clearpage


\section{Problem Setup Details}
\label{app:problem_setup}

Let $x$ denote a prompt (e.g., a math question), and let $y$ denote a sampled solution of length $|y|$ generated autoregressively from a policy $\pi$, i.e., $y \sim \pi(\cdot \mid x)$. We assume a binary reward function $r(x, y) \in \{0, 1\}$, where $r(x, y) = 1$ if the final answer in $y$ is correct and $0$ otherwise. Our goal is to learn a parameterized policy $\pi_\theta$ that maximizes the expected reward over a dataset $\mathcal{D}$ of prompts:
\begin{align}
\label{eq:rl_obj}
    J(\theta) = \mathbb{E}_{x \sim \mathcal{D},\, y \sim \pi_\theta(\cdot \mid x)}[r(x, y)].
\end{align}
Following the standard REINFORCE derivation~\citep{reinforce}, the gradient of this objective can be written as $\nabla_\theta J(\theta) = \mathbb{E}_{x, y} \left[ r(x, y) \nabla_\theta \log \pi_\theta(y \mid x) \right]$.

To reduce the variance of this estimator, it is common to subtract a baseline function that depends only on the prompt $x$, which does not change the optimum of the policy gradient~\citep{Kool2019Buy4R, richter2020vargrad, zhu2023principled, shao2024deepseekmath}. In this work, we use the expected reward under the current policy, $\mathbb{E}_{y' \sim \pi_\theta(\cdot \mid x)}[r(x, y')]$, as the baseline, which is standard in LLM post-training~\citep{shao2024deepseekmath, deepseekai2025deepseekr1, yu2025dapoopensourcellmreinforcement, liu2025understandingr1zeroliketrainingcritical}. Since the reward is binary, we define $p_{\pi_\theta}(x) \coloneqq \mathbb{E}_{y \sim \pi_\theta(\cdot \mid x)}[r(x, y)]$ as the probability of generating a correct answer, and $A(x, y) \coloneqq r(x, y) - p_{\pi_\theta}(x)$ as the advantage.
The policy gradient can be expressed as $\nabla_\theta J(\theta) = \mathbb{E}_{x \sim \mathcal{D},\, y \sim \pi_\theta(\cdot \mid x)} \left[ A(x, y) \nabla_\theta \log \pi_\theta(y \mid x) \right]$.

In practice, LLMs are trained with multiple updates on generations produced by some old policy $\pi_{\theta_\mathrm{old}}$ and the training is often stabilized using techniques such as PPO-style clipping~\citep{schulman2017proximal, shao2024deepseekmath, xiong2025minimalistapproachllmreasoning}. However, we focus on a \textbf{purely on-policy} setting, where each gradient step is followed by the collection of fresh rollouts. Specifically, at each iteration $t$, we perform a single gradient step to maximize:
\begin{align}
    J(\theta) = \mathbb{E}_{x \sim \mathcal{D},\, y \sim \pi_{\theta}(\cdot \mid x)} [ A(x, y) \log{\pi_\theta(y \mid x)} ].
\end{align}
Note that the above objective has the same gradient as:
\begin{align}
\label{eq:pg_adv_x}
    J(\theta) = \mathbb{E}_{x \sim \mathcal{D},\, y \sim \pi_{\theta_t}(\cdot \mid x)} [ A(x, y) \frac{\pi_\theta(y \mid x)}{\pi_{\theta_t}(y \mid x)} ],
\end{align}
since we are purely on-policy and $\pi_{\theta_t}$ is the policy before the update and also serves as the sampling distribution.

Given the autoregressive nature of LLMs, we further decompose the objective into a token-level form, treating each token as an individual action:
\begin{align}
    J(\theta) &= \mathbb{E}_{x \sim \mathcal{D},\, y \sim \pi_{\theta}(\cdot \mid x)} \left[A(x, y) \log \left( \prod_{l=1}^{|y|} \pi_\theta(y_l \mid x, y_{<l}) \right) \right] \\
    &=\mathbb{E}_{x \sim \mathcal{D},\, y \sim \pi_{\theta}(\cdot \mid x)} \left[A(x, y) \sum_{l=1}^{|y|} \log \pi_\theta(y_l \mid x, y_{<l}) \right],
\end{align}
where $y_l$ denotes the $l$-th token in the generated sequence.
Similarly, the above objective has the same gradient as:
\begin{align}
    J(\theta) = \mathbb{E}_{x \sim \mathcal{D},\, y \sim \pi_{\theta_t}(\cdot \mid x)} \left[ A(x, y) \sum_{l=1}^{|y|} \frac{\pi_\theta(y_l \mid x, y_{<l})}{\pi_{\theta_t}(y_l \mid x, y_{<l})} \right].
\end{align}
Normalize by the length of $y$, we arrive at
\begin{align}
\label{eq:token_level_obj}
    J(\theta) = \mathbb{E}_{x \sim \mathcal{D},\, y \sim \pi_{\theta_t}(\cdot \mid x)} \left[ \frac{1}{|y|} A(x, y) \sum_{l=1}^{|y|} \frac{\pi_\theta(y_l \mid x, y_{<l})}{\pi_{\theta_t}(y_l \mid x, y_{<l})} \right].
\end{align}
This objective corresponds to a purely on-policy variant of GRPO~\citep{shao2024deepseekmath, deepseekai2025deepseekr1}, without KL regularization to a fixed reference policy $\pi_{\mathrm{ref}}$~\citep{yu2025dapoopensourcellmreinforcement} and without standard deviation-based advantage regularization~\citep{liu2025understandingr1zeroliketrainingcritical}. We adopt this formulation to eliminate the off-policyness during updates, clipping heuristics, and additional hyperparameters. This results in a \textbf{clean} experimental setup that is directly derived from the original RL objective in Eq.~\ref{eq:rl_obj}.

\clearpage
\section{Preliminary Investigation Details}
\label{app:exp_detail}

\subsection{Dataset Details}
\label{app:data_detail}

{\renewcommand{\arraystretch}{1.1}
\begin{table}[th]\centering
\caption{Dataset split, maximum prompt length, and maximum generation length\label{tab:data_detail}}
\resizebox{1\linewidth}{!}{
\begin{tabular}{ccccc} 
\midrule[0.15ex]
Dataset & Huggingface Dataset Card & Train - Val & Prompt Length & Generation Length \\  \midrule[0.05ex]
MATH & DigitalLearningGmbH/MATH-lighteval & 7.5k - 5k & $1{,}024$ & $4{,}096$ \\
DeepScaleR & agentica-org/DeepScaleR-Preview-Dataset & 40.3k - / & $1{,}024$ & $4{,}096$ \\
\midrule[0.15ex]
\end{tabular}
}
\end{table}
}

{\renewcommand{\arraystretch}{1.1}
\begin{table}[th]\centering
\caption{Model prompt format\label{tab:data_prompt}}
\resizebox{1\linewidth}{!}{
\begin{tabular}{cp{1\linewidth}} 
\midrule[0.15ex]
Model Family & Prompt Format \\  \midrule[0.05ex]
Qwen (Base) & \textbf{\{prompt\}} Let's think step by step and output the final answer within \textbackslash boxed\{\}. \\
Llama (Instruct) & \textless\textbar begin\_of\_text\textbar\textgreater
\textless\textbar start\_header\_id\textbar\textgreater system\textless\textbar end\_header\_id\textbar\textgreater Cutting Knowledge Date: December 2023 Today Date: 26 Jul 2024\textless\textbar eot\_id \textbar\textgreater \textless\textbar  start\_header\_id \textbar\textgreater user \textless\textbar end\_header\_id \textbar\textgreater \textbf{\{prompt\}} Let's think step by step and output the final answer within \textbackslash boxed\{\}. \textless\textbar eot\_id \textbar\textgreater \textless\textbar start\_header\_id \textbar\textgreater assistant \textless\textbar end\_header\_id \textbar\textgreater \\
\midrule[0.15ex]
\end{tabular}
}
\end{table}
}

\subsection{Model Details}

We perform \textbf{full parameter} training on 8 A100 GPUs using Qwen3-1.7B-Base (model card: Qwen/Qwen3-1.7B-Base), Qwen3-4B-Base (model card: Qwen/Qwen3-4B-Base), Qwen3-8B-Base (model card: Qwen/Qwen3-8B-Base), and Llama3.2-3B-it (model card: meta-llama/Llama-3.2-3B-Instruct).

\subsection{Reward Details}

We use a rule-based reward function based on the correctness of the response with math-verify, assigning +1 for correct answers and 0 for incorrect ones or generations that exceed the context length. Recent studies~\citep{chen2025empiricalstudyelicitingimproving} have proposed incorporating format-based rules into reward calculations to encourage models to follow specific output formats. However, in our experiments, we observed no significant difference in performance with or without such format-based rewards. Therefore, for simplicity, we exclude them from our implementation.

\subsection{Evaluation Details}

Following prior work~\citep{zeng2025simplerlzooinvestigatingtamingzero}, we evaluate model performance on a suite of standard mathematical reasoning benchmarks, including MATH500~\citep{hendrycks2021measuringmathematicalproblemsolving}, Minerva Math~\citep{lewkowycz2022solvingquantitativereasoningproblems}, and OlympiadBench~\citep{he2024olympiadbenchchallengingbenchmarkpromoting}, as well as competition-level benchmarks such as AMC 2023, AIME 2024, and AIME 2025.

For smaller-scale datasets, we report results using the average reward across multiple generations. Specifically, for Minerva Math, we report \texttt{Avg@4}; for AMC 2023, AIME 2024, and AIME 2025, we report \texttt{Avg@32}.

For MATH experiments, we use decoding parameters \texttt{top\_k} = 20, \texttt{temperature} = 0.6, and \texttt{top\_p} = 0.95. For DeepScaleR experiments, we use \texttt{top\_k} = $-1$ (i.e., disabled), \texttt{temperature} = 0.6, and \texttt{top\_p} = 0.95.

\newpage
\subsection{Complete List of Experiments}
\label{app:hyper_detail}

The learning rate for each batch size is tuned on a logarithmic scale using the Qwen3-8B-Base model. For all other models, we adopt the corresponding optimal learning rate found for Qwen3-8B-Base. The complete list of all the experiments is provided below with the chosen learning rate highlighted in \textbf{bold}.

\begin{table}[ht]
\centering
\caption{Complete List of Experiments for Math}
\label{tab:all_exp_math}
\resizebox{\textwidth}{!}{%
\begin{tabular}{cccccccc}
\toprule
Model & \#Prompts ($m$) & \#Generations ($n$) & Context Length & Num Workers & Engine & Batch Size ($b$) & LR \\
\midrule
\multirow{7}{*}{Qwen3-8B-base}
& 64 & 16 & 4096 & 8 & VLLM & 1024 & 1E-6/\textbf{2E-6} \\
& 128 & 16 & 4096 & 8 & VLLM & 2048 & 1E-6/\textbf{2E-6}/5E-6/1E-5 \\
& 256 & 16 & 4096 & 8 & VLLM & 4096 & 2E-6/\textbf{4E-6}/8E-6 \\
& 512 & 16 & 4096 & 8 & VLLM & 8192 & 4E-6/\textbf{8E-6}/1.6E-5 \\
& 1024 & 16 & 4096 & 8 & VLLM & 16384 & 4E-6/\textbf{8E-6}/1.6E-5 \\
& 2048 & 16 & 4096 & 8 & VLLM & 32768 & 4E-6/8E-6/\textbf{1.6E-5}/3.2E-5 \\
& 4096 & 16 & 4096 & 8 & VLLM & 65536 & 8E-6/1.6E-5/\textbf{3.2E-5}/6.4E-5 \\

\midrule

\multirow{7}{*}{Qwen3-4B-base}
& 64 & 16 & 4096 & 8 & VLLM & 1024 & 2.00E-06 \\
& 128 & 16 & 4096 & 8 & VLLM & 2048 & 2.00E-06 \\
& 256 & 16 & 4096 & 8 & VLLM & 4096 & 4.00E-06 \\
& 512 & 16 & 4096 & 8 & VLLM & 8192 & 8.00E-06 \\
& 1024 & 16 & 4096 & 8 & VLLM & 16384 & 8.00E-06 \\
& 2048 & 16 & 4096 & 8 & VLLM & 32768 & 1.60E-05 \\
& 4096 & 16 & 4096 & 8 & VLLM & 65536 & 3.20E-05 \\

\midrule

\multirow{7}{*}{Qwen3-1.7B-base}
& 64 & 16 & 4096 & 8 & VLLM & 1024 & 2.00E-06 \\
& 128 & 16 & 4096 & 8 & VLLM & 2048 & 2.00E-06 \\
& 256 & 16 & 4096 & 8 & VLLM & 4096 & 4.00E-06 \\
& 512 & 16 & 4096 & 8 & VLLM & 8192 & 8.00E-06 \\
& 1024 & 16 & 4096 & 8 & VLLM & 16384 & 8.00E-06 \\
& 2048 & 16 & 4096 & 8 & VLLM & 32768 & 1.60E-05 \\
& 4096 & 16 & 4096 & 8 & VLLM & 65536 & 3.20E-05 \\

\midrule

\multirow{7}{*}{Llama3.2-3B-it}
& 64 & 16 & 4096 & 8 & VLLM & 1024 & 2.00E-06 \\
& 128 & 16 & 4096 & 8 & VLLM & 2048 & 2.00E-06 \\
& 256 & 16 & 4096 & 8 & VLLM & 4096 & 4.00E-06 \\
& 512 & 16 & 4096 & 8 & VLLM & 8192 & 8.00E-06 \\
& 1024 & 16 & 4096 & 8 & VLLM & 16384 & 8.00E-06 \\
& 2048 & 16 & 4096 & 8 & VLLM & 32768 & 8.00E-06 \\
& 4096 & 16 & 4096 & 8 & VLLM & 65536 & 1.20E-05 \\

\midrule

\multirow{6}{*}{Qwen3-4B-base}
& 32 & 32 & 4096 & 8 & VLLM & 1024 & 2.00E-06 \\
& 256 & 32 & 4096 & 8 & VLLM & 8192 & 8.00E-06 \\
& 2048 & 32 & 4096 & 8 & VLLM & 65536 & 3.20E-05 \\
& 16 & 64 & 4096 & 8 & VLLM & 1024 & 2.00E-06 \\
& 128 & 64 & 4096 & 8 & VLLM & 8192 & 8.00E-06 \\
& 1024 & 64 & 4096 & 8 & VLLM & 65536 & 3.20E-05 \\

\bottomrule
\end{tabular}
}
\end{table}

\begin{table}[ht]
\centering
\caption{Complete List of Experiments for DeepScaleR}
\label{tab:all_exp_deepscaler}
\resizebox{\textwidth}{!}{%
\begin{tabular}{cccccccc}
\toprule
Model & \#Prompts ($m$) & \#Generations ($n$) & Context Length & Num Workers & Engine & Batch Size & LR \\
\midrule

\multirow{7}{*}{Qwen3-8B-base}
& 64 & 16 & 4096 & 8 & VLLM & 1024 & 1E-6/\textbf{2E-6} \\
& 128 & 16 & 4096 & 8 & VLLM & 2048 & 1E-6/\textbf{2E-6}/5E-6/1E-5/2E-5 \\
& 256 & 16 & 4096 & 8 & VLLM & 4096 & 2E-6/\textbf{4E-6}/8E-6 \\
& 512 & 16 & 4096 & 8 & VLLM & 8192 & 2E-6/\textbf{4E-6}/6E-6 \\
& 1024 & 16 & 4096 & 8 & VLLM & 16384 & 4E-6/\textbf{8E-6}/1.2E-5/1.6E-5 \\
& 2048 & 16 & 4096 & 8 & VLLM & 32768 & 8E-6/\textbf{1.2E-5}/1.6E-5 \\
& 4096 & 16 & 4096 & 8 & VLLM & 65536 & 8E-6/\textbf{1.2E-5}/1.6E-5 \\

\midrule

\multirow{7}{*}{Qwen3-4B-base}
& 64 & 16 & 4096 & 8 & VLLM & 1024 & 2.00E-06 \\
& 128 & 16 & 4096 & 8 & VLLM & 2048 & 2.00E-06 \\
& 256 & 16 & 4096 & 8 & VLLM & 4096 & 4.00E-06 \\
& 512 & 16 & 4096 & 8 & VLLM & 8192 & 4.00E-06 \\
& 1024 & 16 & 4096 & 8 & VLLM & 16384 & 8.00E-06 \\
& 2048 & 16 & 4096 & 8 & VLLM & 32768 & 1.20E-05 \\
& 4096 & 16 & 4096 & 8 & VLLM & 65536 & 1.20E-05 \\

\midrule

\multirow{7}{*}{Qwen3-1.7B-base}
& 64 & 16 & 4096 & 8 & VLLM & 1024 & 2.00E-06 \\
& 128 & 16 & 4096 & 8 & VLLM & 2048 & 2.00E-06 \\
& 256 & 16 & 4096 & 8 & VLLM & 4096 & 4.00E-06 \\
& 512 & 16 & 4096 & 8 & VLLM & 8192 & 4.00E-06 \\
& 1024 & 16 & 4096 & 8 & VLLM & 16384 & 8.00E-06 \\
& 2048 & 16 & 4096 & 8 & VLLM & 32768 & 1.20E-05 \\
& 4096 & 16 & 4096 & 8 & VLLM & 65536 & 1.20E-05 \\

\midrule

\multirow{7}{*}{Llama3.2-3B-it}
& 64 & 16 & 4096 & 8 & VLLM & 1024 & 2.00E-06 \\
& 128 & 16 & 4096 & 8 & VLLM & 2048 & 2.00E-06 \\
& 256 & 16 & 4096 & 8 & VLLM & 4096 & 4.00E-06 \\
& 512 & 16 & 4096 & 8 & VLLM & 8192 & 4.00E-06 \\
& 1024 & 16 & 4096 & 8 & VLLM & 16384 & 8.00E-06 \\
& 2048 & 16 & 4096 & 8 & VLLM & 32768 & 8.00E-06 \\
& 4096 & 16 & 4096 & 8 & VLLM & 65536 & 1.20E-05 \\

\midrule

\multirow{6}{*}{Qwen3-4B-base}
& 32 & 32 & 4096 & 8 & VLLM & 1024 & 2.00E-06 \\
& 256 & 32 & 4096 & 8 & VLLM & 8192 & 4.00E-06 \\
& 2048 & 32 & 4096 & 8 & VLLM & 65536 & 1.20E-05 \\
& 16 & 64 & 4096 & 8 & VLLM & 1024 & 2.00E-06 \\
& 128 & 64 & 4096 & 8 & VLLM & 8192 & 4.00E-06 \\
& 1024 & 64 & 4096 & 8 & VLLM & 65536 & 1.20E-05 \\

\midrule

\multirow{7}{*}{Qwen3-4B-base}
& 64 & 16 & 8192 & 8 & VLLM & 1024 & 2.00E-06 \\
& 128 & 16 & 8192 & 8 & VLLM & 2048 & 2.00E-06 \\
& 256 & 16 & 8192 & 8 & VLLM & 4096 & 4.00E-06 \\
& 512 & 16 & 8192 & 8 & VLLM & 8192 & 4.00E-06 \\
& 1024 & 16 & 8192 & 8 & VLLM & 16384 & 8.00E-06 \\
& 2048 & 16 & 8192 & 8 & VLLM & 32768 & 1.20E-05 \\
& 4096 & 16 & 8192 & 8 & VLLM & 65536 & 1.20E-05 \\

\midrule

\multirow{7}{*}{Qwen3-4B-base}
& 16 & 16 & 4096 & 1 & VLLM & 256 & 1.00E-06 \\
& 32 & 16 & 4096 & 1 & VLLM & 512 & 1.00E-06 \\
& 64 & 16 & 4096 & 1 & VLLM & 1024 & 2.00E-06 \\
& 128 & 16 & 4096 & 1 & VLLM & 2048 & 2.00E-06 \\
& 256 & 16 & 4096 & 1 & VLLM & 4096 & 4.00E-06 \\
& 512 & 16 & 4096 & 1 & VLLM & 8192 & 4.00E-06 \\
& 1024 & 16 & 4096 & 1 & VLLM & 16384 & 8.00E-06 \\

\midrule

\multirow{9}{*}{Qwen3-4B-base}
& 16 & 16 & 4096 & 8 & SGLang & 256 & 1.00E-06 \\
& 32 & 16 & 4096 & 8 & SGLang & 512 & 1.00E-06 \\
& 64 & 16 & 4096 & 8 & SGLang & 1024 & 2.00E-06 \\
& 128 & 16 & 4096 & 8 & SGLang & 2048 & 2.00E-06 \\
& 256 & 16 & 4096 & 8 & SGLang & 4096 & 4.00E-06 \\
& 512 & 16 & 4096 & 8 & SGLang & 8192 & 4.00E-06 \\
& 1024 & 16 & 4096 & 8 & SGLang & 16384 & 8.00E-06 \\
& 2048 & 16 & 4096 & 8 & SGLang & 32768 & 1.20E-05 \\
& 4096 & 16 & 4096 & 8 & SGLang & 65536 & 1.20E-05 \\

\bottomrule
\end{tabular}
}
\end{table}

\begin{table}[ht]
\centering
\caption{Complete List of Experiments for DeepScaleR (cont.)}
\label{tab:all_exp_deepscaler_2}
\resizebox{\textwidth}{!}{%
\begin{tabular}{cccccccc}
\toprule
Model & \#Prompts ($m$) & \#Generations ($n$) & Context Length & Num Workers & Engine & Batch Size & LR \\
\midrule

\multirow{9}{*}{Qwen3-4B-Base \& $p(x)=0$}
& 32 & 128 & 4096 & 8 & VLLM & 4096 & 4.00E-06 \\
& 64 & 64 & 4096 & 8 & VLLM & 4096 & 4.00E-06 \\
& 128 & 32 & 4096 & 8 & VLLM & 4096 & 4.00E-06 \\
& 256 & 16 & 4096 & 8 & VLLM & 4096 & 4.00E-06 \\
& 512 & 8 & 4096 & 8 & VLLM & 4096 & 4.00E-06 \\
& 1024 & 4 & 4096 & 8 & VLLM & 4096 & 4.00E-06 \\
& 2048 & 2 & 4096 & 8 & VLLM & 4096 & 4.00E-06 \\

\midrule

\multirow{9}{*}{Qwen3-4B-Base \& $p(x)=0.25$}
& 32 & 128 & 4096 & 8 & VLLM & 4096 & 4.00E-06 \\
& 64 & 64 & 4096 & 8 & VLLM & 4096 & 4.00E-06 \\
& 128 & 32 & 4096 & 8 & VLLM & 4096 & 4.00E-06 \\
& 256 & 16 & 4096 & 8 & VLLM & 4096 & 4.00E-06 \\
& 512 & 8 & 4096 & 8 & VLLM & 4096 & 4.00E-06 \\
& 1024 & 4 & 4096 & 8 & VLLM & 4096 & 4.00E-06 \\
& 2048 & 2 & 4096 & 8 & VLLM & 4096 & 4.00E-06 \\

\midrule

\multirow{9}{*}{Qwen3-4B-Base \& $p(x)=0.5$}
& 32 & 128 & 4096 & 8 & VLLM & 4096 & 4.00E-06 \\
& 64 & 64 & 4096 & 8 & VLLM & 4096 & 4.00E-06 \\
& 128 & 32 & 4096 & 8 & VLLM & 4096 & 4.00E-06 \\
& 256 & 16 & 4096 & 8 & VLLM & 4096 & 4.00E-06 \\
& 512 & 8 & 4096 & 8 & VLLM & 4096 & 4.00E-06 \\
& 1024 & 4 & 4096 & 8 & VLLM & 4096 & 4.00E-06 \\
& 2048 & 2 & 4096 & 8 & VLLM & 4096 & 4.00E-06 \\

\midrule

\multirow{9}{*}{Qwen3-4B-Base \& $p(x)=0.75$}
& 32 & 128 & 4096 & 8 & VLLM & 4096 & 4.00E-06 \\
& 64 & 64 & 4096 & 8 & VLLM & 4096 & 4.00E-06 \\
& 128 & 32 & 4096 & 8 & VLLM & 4096 & 4.00E-06 \\
& 256 & 16 & 4096 & 8 & VLLM & 4096 & 4.00E-06 \\
& 512 & 8 & 4096 & 8 & VLLM & 4096 & 4.00E-06 \\
& 1024 & 4 & 4096 & 8 & VLLM & 4096 & 4.00E-06 \\
& 2048 & 2 & 4096 & 8 & VLLM & 4096 & 4.00E-06 \\

\midrule

\multirow{9}{*}{Qwen3-4B-Base \& $p(x)=1$}
& 32 & 128 & 4096 & 8 & VLLM & 4096 & 4.00E-06 \\
& 64 & 64 & 4096 & 8 & VLLM & 4096 & 4.00E-06 \\
& 128 & 32 & 4096 & 8 & VLLM & 4096 & 4.00E-06 \\
& 256 & 16 & 4096 & 8 & VLLM & 4096 & 4.00E-06 \\
& 512 & 8 & 4096 & 8 & VLLM & 4096 & 4.00E-06 \\
& 1024 & 4 & 4096 & 8 & VLLM & 4096 & 4.00E-06 \\
& 2048 & 2 & 4096 & 8 & VLLM & 4096 & 4.00E-06 \\

\bottomrule
\end{tabular}
}
\end{table}

\clearpage
\section{Preliminary Investigation Complete Results}
\label{app:complete_results}

\subsection{Complete Results for Section~\ref{sec:optimal_bs}}

\subsubsection{\texorpdfstring{Results with varying $m$}{Results with varying m}}

\begin{figure}[h]
    \centering
    \includegraphics[trim={0 0 0 0}, clip, width=0.74\textwidth]{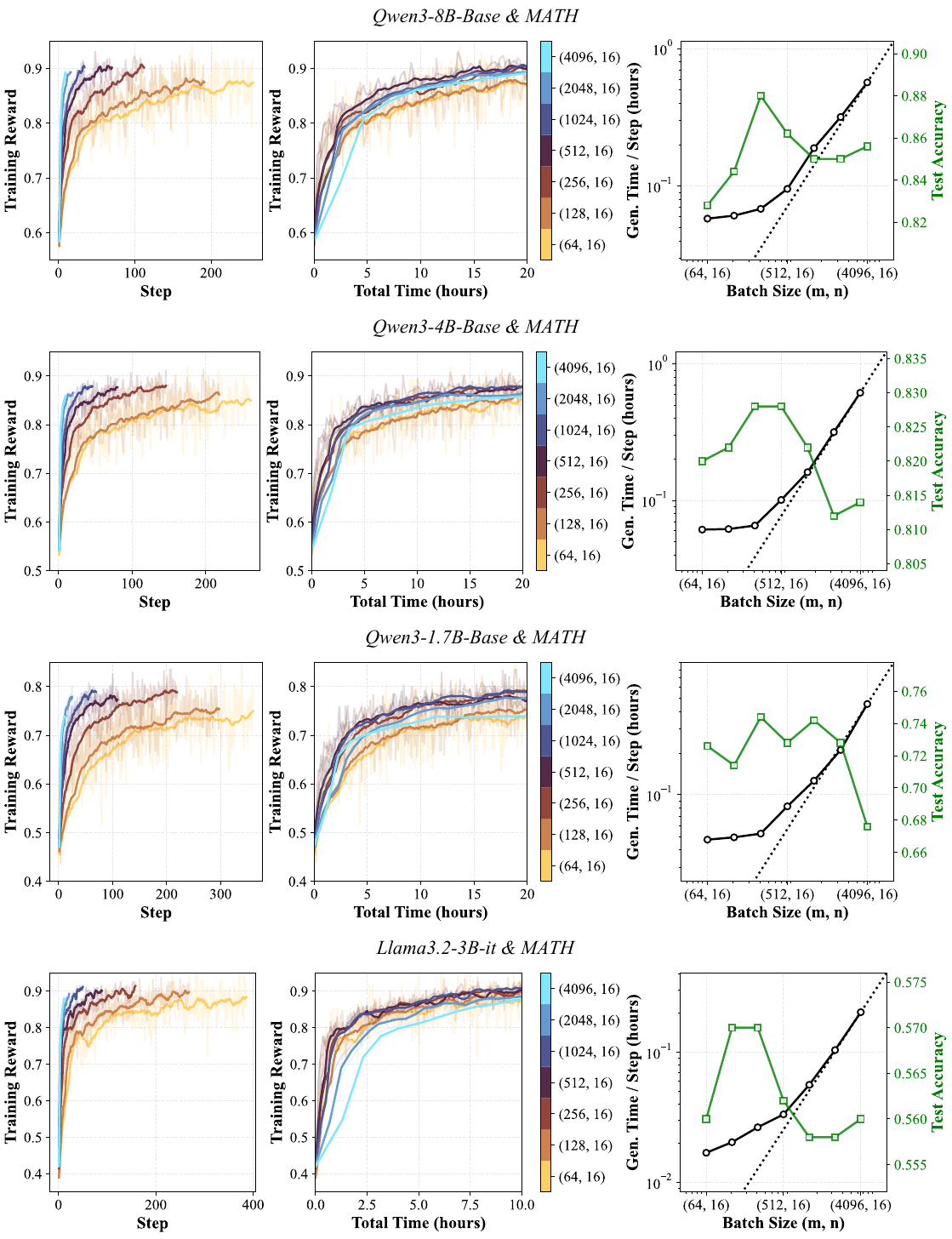}
    \vskip -0.3cm
    \caption{Results for all four models on MATH with $n=16$. (Left / Middle) Training reward as a function of training steps and wall-clock time. The legend indicates the batch configuration in terms of (number of prompts $m$, generations per prompt $n$). (Right) Generation time per step and test accuracy across different batch sizes. The dashed line represents the linear increase that intercepts the origin and the generation time for the largest batch size. Both axes are in log scale.}
\end{figure}

\clearpage
\begin{figure}[ht]
    \centering
    \includegraphics[trim={0 0 0 0}, clip, width=0.74\textwidth]{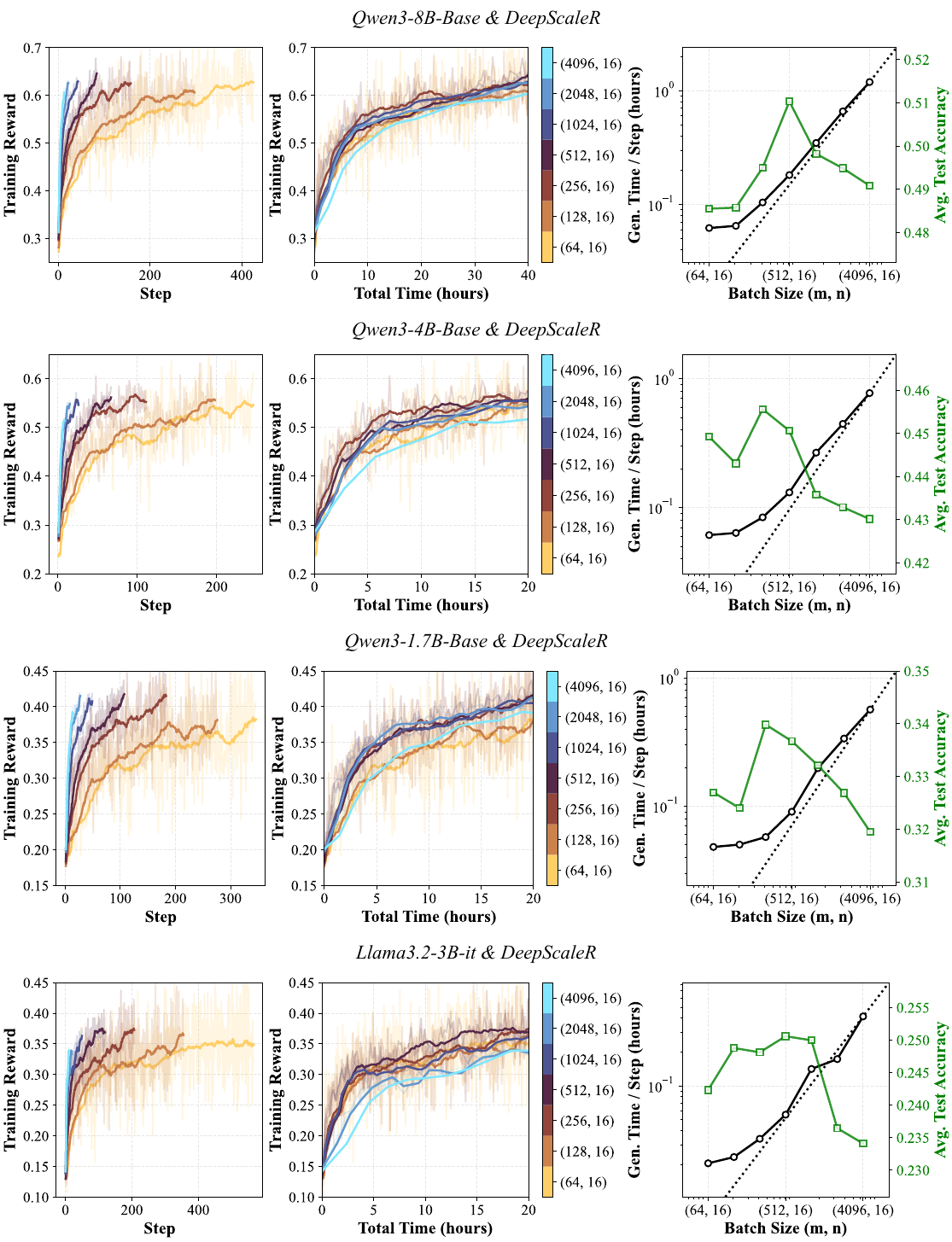}
    \vskip -0.3cm
    \caption{Results for all four models on DeepScaleR with $n=16$. (Left / Middle) Training reward as a function of training steps and wall-clock time. The legend indicates the batch configuration in terms of (number of prompts $m$, generations per prompt $n$). (Right) Generation time per step and test accuracy across different batch sizes. The dashed line represents the linear increase that intercepts the origin and the generation time for the largest batch size. Both axes are in log scale.}
    \label{fig:dp_1}
\end{figure}

\clearpage
\begin{table}[h]\centering
\caption{{Detailed Results for Fig.~\ref{fig:dp_1}.}}
\begin{tabular}[t]{c|c|cccccc|c} 
\midrule[0.15ex]
\multirow{2}{*}{Model} & \multirow{2}{*}{$m$} & \multirow{2}{*}{MATH500} & \multirow{2}{*}{Olymp.} & Minerva & AMC23 & AIME24 & AIME25 & \multirow{2}{*}{Avg.} \\
 & & & & Avg@4 & Avg@32 & Avg@32 & Avg@32 & \\
\midrule[0.05ex]
\multirow{8}{*}{Qwen3-8B-Base}
& $\pi_\mathrm{ref}$ & 70.2 & 34.3 & 29.8 & 49.1 & 15.8 & 8.8 & 34.7 \\
& 4096 & 85.8 & 52.2 & 43.0 & 70.9 & 21.5 & 21.1 & 49.1 \\
& 2048  & 85.2 & 53.4 & 43.9 & 66.6 & 26.4 & 21.5 & 49.5\\
& 1024 & 85.6 & 54.9 & 45.9 & 70.5 & 22.8 & 19.3 & 49.8 \\
& 512 & 87.2 & 55.8 & 44.1 & 71.8 & 26.1 & 21.1 & 51.0 \\
& 256 & 85.0 & 57.4 & 40.4 & 66.3 & 24.9 & 22.9 & 49.5 \\
& 128 & 85.6 & 53.9 & 42.0 & 67.8 & 22.3 & 19.9 & 48.6 \\
& 64 & 85.2 & 54.7 & 42.1 & 70.6 & 21.4 & 17.3 & 48.6 \\
\midrule[0.05ex]
\multirow{8}{*}{Qwen3-4B-Base}
& $\pi_\mathrm{ref}$ & 65.8 & 34.4 & 26.9 & 47.3 & 10.9 & 7.1 & 32.1\\
& 4096 & 80.6 & 45.8 & 39.7 & 59.8 & 16.4 & 15.8 & 43.0 \\
& 2048 & 83.2 & 48.4 & 39.2 & 57.0 & 16.0 & 15.9 & 43.3 \\
& 1024 & 81.6 & 46.1 & 40.2 & 59.3 & 18.1 & 16.1 & 43.6 \\
& 512 & 84.0 & 49.7 & 38.8 & 62.9 & 17.1 & 17.9 & 45.1 \\
& 256 & 82.8 & 48.1 & 40.3 & 66.3 & 17.8 & 18.1 & 45.6 \\
& 128 & 83.8 & 46.0 & 42.6 & 59.5 & 18.2 & 15.6 & 44.3 \\
& 64 & 83.2 & 48.2 & 39.7 & 64.1 & 17.6 & 16.8 & 44.9 \\
\midrule[0.05ex]
\multirow{8}{*}{Qwen3-1.7B-Base}
& $\pi_\mathrm{ref}$ & 57.0 & 23.9 & 21.8 & 29.0 & 3.8 & 1.1 & 22.8 \\
& 4096 & 69.8 & 35.2 & 29.0 & 40.7 & 9.1 & 8.0 & 32.0 \\
& 2048 & 70.2 & 34.3 & 31.2 & 42.0 & 12.2 & 6.2 & 32.7 \\
& 1024 & 72.2 & 36.2 & 29.7 & 41.8 & 12.4 & 7.0 & 33.2 \\
& 512 & 71.8 & 37.1 & 30.1 & 44.2 & 12.7 & 6.1 & 33.7 \\
& 256 & 72.6 & 35.6 & 31.5 & 46.9 & 10.1 & 7.2 & 34.0 \\
& 128 & 68.4 & 35.2 & 30.0 & 43.3 & 10.9 & 6.7 & 32.4 \\
& 64 & 70.2 & 36.5 & 30.0 & 40.8 & 11.2 & 7.5 & 32.7 \\
\midrule[0.05ex]
\multirow{8}{*}{Llama3.2-3B-it}
& $\pi_\mathrm{ref}$ & 42.8 & 12.3 & 13.8 & 19.7 & 4.6 & 0.4 & 15.6 \\
& 4096 & 55.2 & 20.0 & 21.1 & 31.6 & 11.9 & 0.6 & 23.4 \\
& 2048 & 55.8 & 19.3 & 21.2 & 30.9 & 13.8 & 0.9 & 23.6 \\
& 1024 & 57.8 & 22.8 & 21.2 & 34.8 & 12.1 & 1.2 & 25.0 \\
& 512 & 58.0 & 22.3 & 22.7 & 30.0 & 15.8 & 1.6 & 25.1 \\
& 256 & 57.6 & 21.8 & 22.6 & 32.0 & 14.5 & 0.4 & 24.8 \\
& 128 & 55.6 & 22.7 & 22.2 & 34.8 & 13.9 & 0.1 & 24.9 \\
& 64 & 56.8 & 20.9 & 25.5 & 31.8 & 10.2 & 0.2 & 24.2 \\
\midrule[0.15ex]
\end{tabular}
\end{table}

\subsubsection{\texorpdfstring{Results with varying $m$ and $n$}{Results with varying m and n}}
\begin{table}[hbt!]\centering
\caption{{Detailed DeepScaleR Results for Fig.~\ref{fig:dp_2}.}}
\begin{tabular}[t]{c|cc|cccccc|c} 
\midrule[0.15ex]
\multirow{2}{*}{Model} & \multirow{2}{*}{$m$} & \multirow{2}{*}{$n$} & \multirow{2}{*}{MATH500} & \multirow{2}{*}{Olymp.} & Minerva & AMC23 & AIME24 & AIME25 & \multirow{2}{*}{Avg.} \\
 & & & & & Avg@4 & Avg@32 & Avg@32 & Avg@32 & \\
\midrule[0.05ex]
\multirow{8}{*}{Qwen3-4B-Base}
& $\pi_\mathrm{ref}$ & \ & 65.8 & 34.4 & 26.9 & 47.3 & 10.9 & 7.1 & 32.1\\
& 32 & 32 & 80.4 & 47.5 & 37.4 & 57.4 & 17.5 & 14.4 & 42.4 \\
& 256 & 32 & 83.2 & 49.1 & 38.6 & 63.4 & 17.3 & 16.0 & 44.6 \\
& 2048 & 32 & 81.4 & 46.3 & 39.8 & 57.6 & 17.6 & 15.5 & 43.0 \\
& 16 & 64 & 81.6 & 47.8 & 39.2 & 58.4 & 15.7 & 14.3 & 42.8 \\
& 128 & 64 & 83.4 & 44.4 & 41.5 & 60.0 & 17.0 & 13.1 & 43.2 \\
& 1024 & 64 & 80.8 & 48.7 & 39.2 & 57.0 & 16.4 & 13.8 & 42.6 \\
\midrule[0.15ex]
\end{tabular}
\end{table}

\clearpage
\begin{figure}[h]
    \centering
    \includegraphics[trim={0 0 0 0}, clip, width=0.74\textwidth]{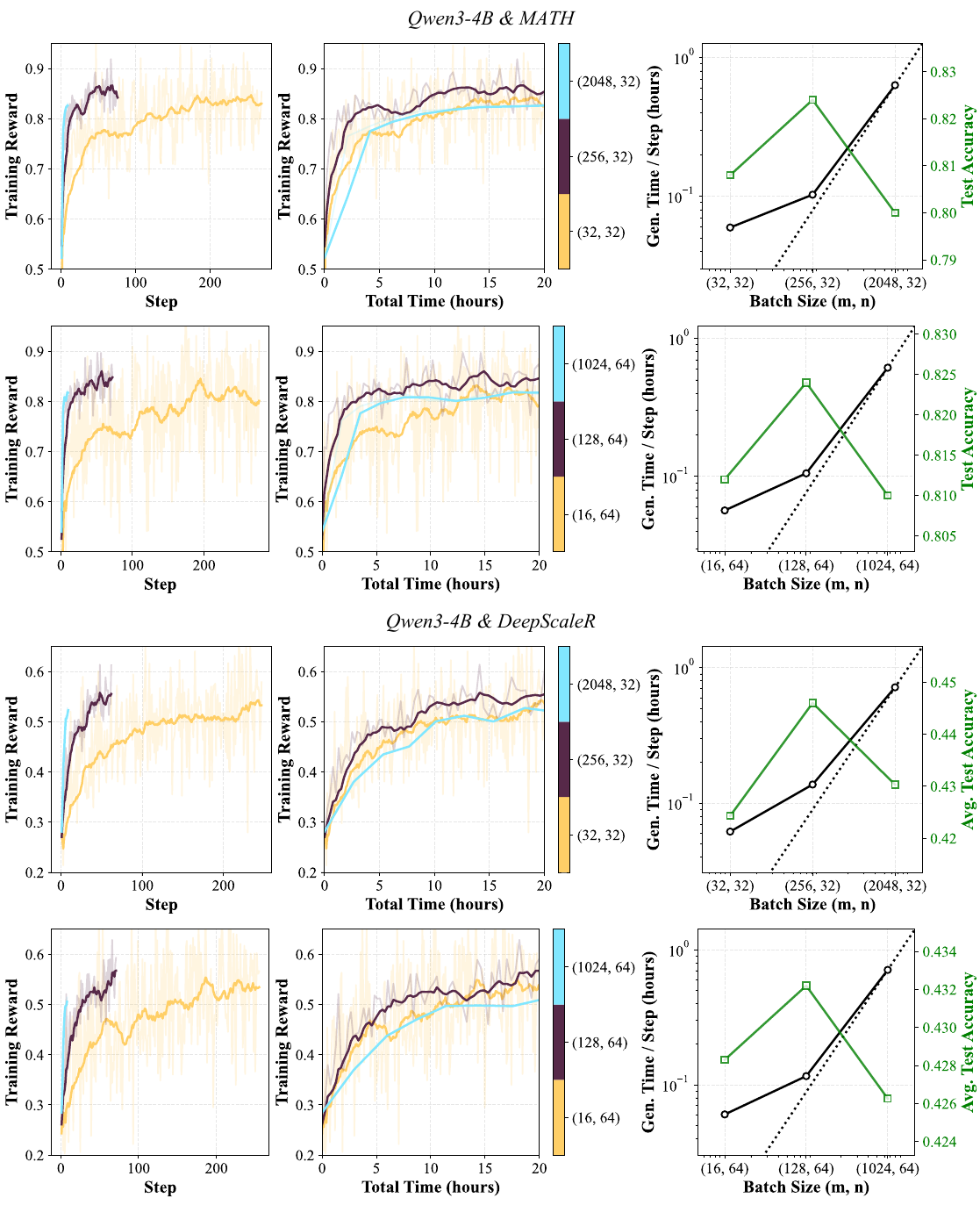}
    \vskip -0.3cm
    \caption{Results for Qwen3-4B on MATH and DeepScaleR with $n=32$ and $64$. (Left / Middle) Training reward as a function of training steps and wall-clock time. The legend indicates the batch configuration in terms of (number of prompts $m$, generations per prompt $n$). (Right) Generation time per step and test accuracy across different batch sizes. The dashed line represents the linear increase that intercepts the origin and the generation time for the largest batch size. Both axes are in log scale.}
    \label{fig:dp_2}
\end{figure}

\clearpage
\subsubsection{Results with a different context length}

\begin{figure}[h]
    \centering
    \includegraphics[trim={0 0 0 0}, clip, width=\textwidth]{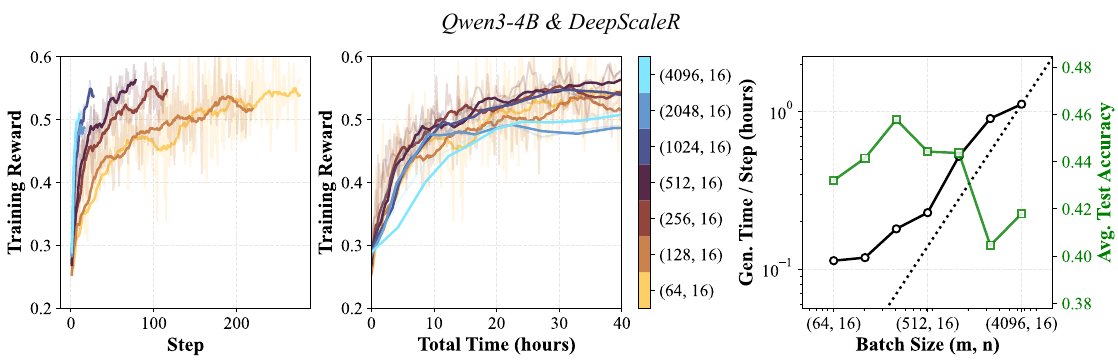}
    \vskip -0.3cm
    \caption{Results for Qwen3-4B on DeepScaleR with context length 8192 (other results are using 4096 context length). (Left / Middle) Training reward as a function of training steps and wall-clock time. The legend indicates the batch configuration in terms of (number of prompts $m$, generations per prompt $n$). (Right) Generation time per step and test accuracy across different batch sizes. The dashed line represents the linear increase that intercepts the origin and the generation time for the largest batch size. Both axes are in log scale.}
    \label{fig:dp_3}
\end{figure}

\begin{table}[h]\centering
\caption{{Detailed DeepScaleR Results for Fig.~\ref{fig:dp_3}.}}
\resizebox{\linewidth}{!}{
\begin{tabular}[t]{c|cc|cccccc|c} 
\midrule[0.15ex]
\multirow{2}{*}{Model} & \multirow{2}{*}{$m$} & \multirow{2}{*}{Context Len.} & \multirow{2}{*}{MATH500} & \multirow{2}{*}{Olymp.} & Minerva & AMC23 & AIME24 & AIME25 & \multirow{2}{*}{Avg.} \\
 & & & & & Avg@4 & Avg@32 & Avg@32 & Avg@32 & \\
\midrule[0.05ex]
\multirow{8}{*}{Qwen3-4B-Base}
& $\pi_\mathrm{ref}$ & \ & 65.8 & 34.4 & 26.9 & 47.3 & 10.9 & 7.1 & 32.1\\
& 64 & 8K & 80.4 & 47.9 & 39.2 & 60.7 & 15.9 & 15.1 & 43.2 \\
& 128 & 8K & 83.0 & 50.6 & 38.4 & 62.3 & 14.2 & 16.4 & 44.1 \\
& 256 & 8K & 82.8 & 50.4 & 42.2 & 62.0 & 19.3 & 18.0 & 45.8 \\
& 512 & 8K & 81.2 & 49.6 & 41.0 & 65.3 & 17.0 & 12.5 & 44.4 \\
& 1024 & 8K & 80.8 & 46.6 & 39.7 & 63.7 & 16.6 & 18.9 & 44.4 \\
& 2048 & 8K & 77.6 & 44.4 & 37.5 & 54.5 & 14.2 & 14.6 & 40.5 \\
& 4096 & 8K & 79.8 & 45.3 & 38.9 & 58.4 & 14.7 & 13.8 & 41.8 \\
\midrule[0.15ex]
\end{tabular}
}
\end{table}

\clearpage
\subsubsection{Results with a different hardware configuration}

\begin{figure}[hbt]
    \centering
    \includegraphics[trim={0 0 0 0}, clip, width=\textwidth]{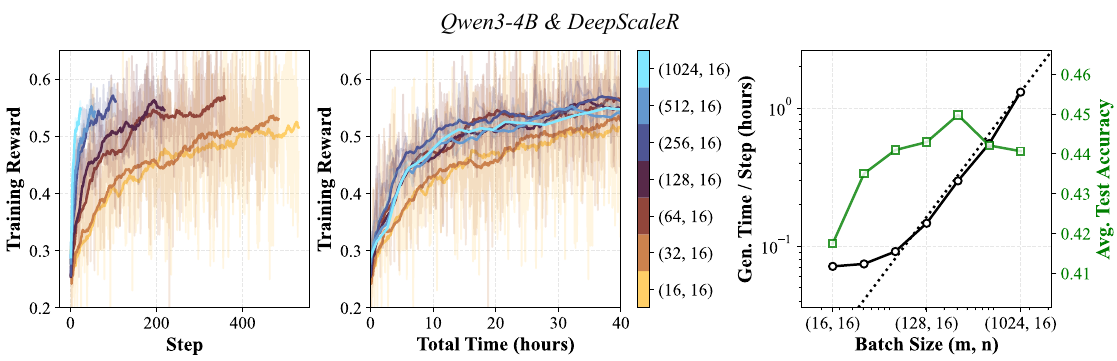}
    \vskip -0.3cm
    \caption{Results for Qwen3-4B on DeepScaleR with only 1 rollout worker with 8 GPUs (other results are using 8 rollout workers, 1 per GPU). (Left / Middle) Training reward as a function of training steps and wall-clock time. The legend indicates the batch configuration in terms of (number of prompts $m$, generations per prompt $n$). (Right) Generation time per step and test accuracy across different batch sizes. The dashed line represents the linear increase that intercepts the origin and the generation time for the largest batch size. Both axes are in log scale.}
    \label{fig:dp_4}
\end{figure}

\begin{table}[hbt]\centering
\caption{{Detailed DeepScaleR Results for Fig.~\ref{fig:dp_4}.}}
\resizebox{\linewidth}{!}{
\begin{tabular}[t]{c|cc|cccccc|c} 
\midrule[0.15ex]
\multirow{2}{*}{Model} & \multirow{2}{*}{$m$} & \multirow{2}{*}{Num. Worker} & \multirow{2}{*}{MATH500} & \multirow{2}{*}{Olymp.} & Minerva & AMC23 & AIME24 & AIME25 & \multirow{2}{*}{Avg.} \\
 & & & & & Avg@4 & Avg@32 & Avg@32 & Avg@32 & \\
\midrule[0.05ex]
\multirow{8}{*}{Qwen3-4B-Base}
& $\pi_\mathrm{ref}$ & \ & 65.8 & 34.4 & 26.9 & 47.3 & 10.9 & 7.1 & 32.1\\
& 16 & 1 & 78.4 & 47.2 & 39.2 & 58.0 & 15.3 & 12.4 & 41.7 \\
& 32 & 1 & 81.8 & 47.0 & 39.0 & 61.3 & 17.1 & 14.9 & 43.5 \\
& 64 & 1 & 83.6 & 49.4 & 40.6 & 58.7 & 15.9 & 16.4 & 44.1 \\
& 128 & 1 & 82.4 & 48.7 & 39.2 & 62.7 & 17.2 & 15.7 & 44.3 \\
& 256 & 1 & 82.8 & 49.6 & 38.6 & 62.6 & 18.9 & 17.5 & 45.0 \\
& 512 & 1 & 81.6 & 47.6 & 40.7 & 62.1 & 17.7 & 15.5 & 44.2 \\
& 1024 & 1 & 80.4 & 47.8 & 40.4 & 60.1 & 18.9 & 16.9 & 44.1 \\
\midrule[0.15ex]
\end{tabular}
}
\end{table}

\clearpage
\subsubsection{Results with a different inference engine}

\begin{figure}[ht]
    \centering
    \includegraphics[trim={0 0 0 0}, clip, width=\textwidth]{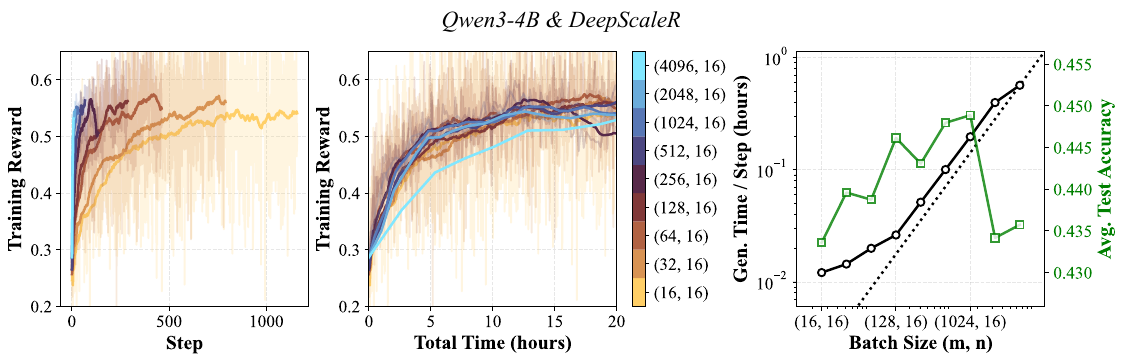}
    \vskip -0.3cm
    \caption{Results for Qwen3-4B on DeepScaleR with SGLang (other results are using VLLM). (Left / Middle) Training reward as a function of training steps and wall-clock time. The legend indicates the batch configuration in terms of (number of prompts $m$, generations per prompt $n$). (Right) Generation time per step and test accuracy across different batch sizes. The dashed line represents the linear increase that intercepts the origin and the generation time for the largest batch size. Both axes are in log scale.}
    \label{fig:dp_5}
\end{figure}

\begin{table}[ht]\centering
\caption{{Detailed DeepScaleR Results for Fig.~\ref{fig:dp_5}.}}
\resizebox{\linewidth}{!}{
\begin{tabular}[t]{c|cc|cccccc|c} 
\midrule[0.15ex]
\multirow{2}{*}{Model} & \multirow{2}{*}{$m$} & \multirow{2}{*}{Inference Eng.} & \multirow{2}{*}{MATH500} & \multirow{2}{*}{Olymp.} & Minerva & AMC23 & AIME24 & AIME25 & \multirow{2}{*}{Avg.} \\
 & & & & & Avg@4 & Avg@32 & Avg@32 & Avg@32 & \\
\midrule[0.05ex]
\multirow{8}{*}{Qwen3-4B-Base}
& $\pi_\mathrm{ref}$ & \ & 65.8 & 34.4 & 26.9 & 47.3 & 10.9 & 7.1 & 32.1\\
& 16 & SGLang & 81.8 & 48.5 & 37.3 & 58.0 & 17.1 & 17.4 & 43.4 \\
& 32 & SGLang & 81.2 & 49.9 & 40.0 & 60.9 & 17.0 & 14.9 & 44.0 \\
& 64 & SGLang & 81.6 & 50.9 & 39.8 & 60.2 & 15.3 & 15.4 & 43.9 \\
& 128 & SGLang & 84.0 & 48.8 & 39.7 & 62.9 & 15.2 & 17.1 & 44.6 \\
& 256 & SGLang & 81.8 & 49.4 & 40.0 & 60.4 & 16.6 & 17.7 & 44.3 \\
& 512 & SGLang & 82.2 & 50.3 & 40.9 & 63.2 & 17.0 & 15.2 & 44.8 \\
& 1024 & SGLang & 81.2 & 51.3 & 41.0 & 62.9 & 16.8 & 16.1 & 44.9 \\
& 2048 & SGLang & 81.2 & 46.0 & 40.6 & 61.0 & 17.3 & 14.4 & 43.4 \\
& 4096 & SGLang & 82.0 & 47.8 & 39.8 & 61.9 & 16.2 & 13.8 & 43.6 \\
\midrule[0.15ex]
\end{tabular}
}
\end{table}

\clearpage
\subsection{Complete Results for Section~\ref{sec:optimal_nm}}

\begin{figure}[ht]
    \centering
    \includegraphics[trim={0 0 0 0}, clip, width=0.7\textwidth]{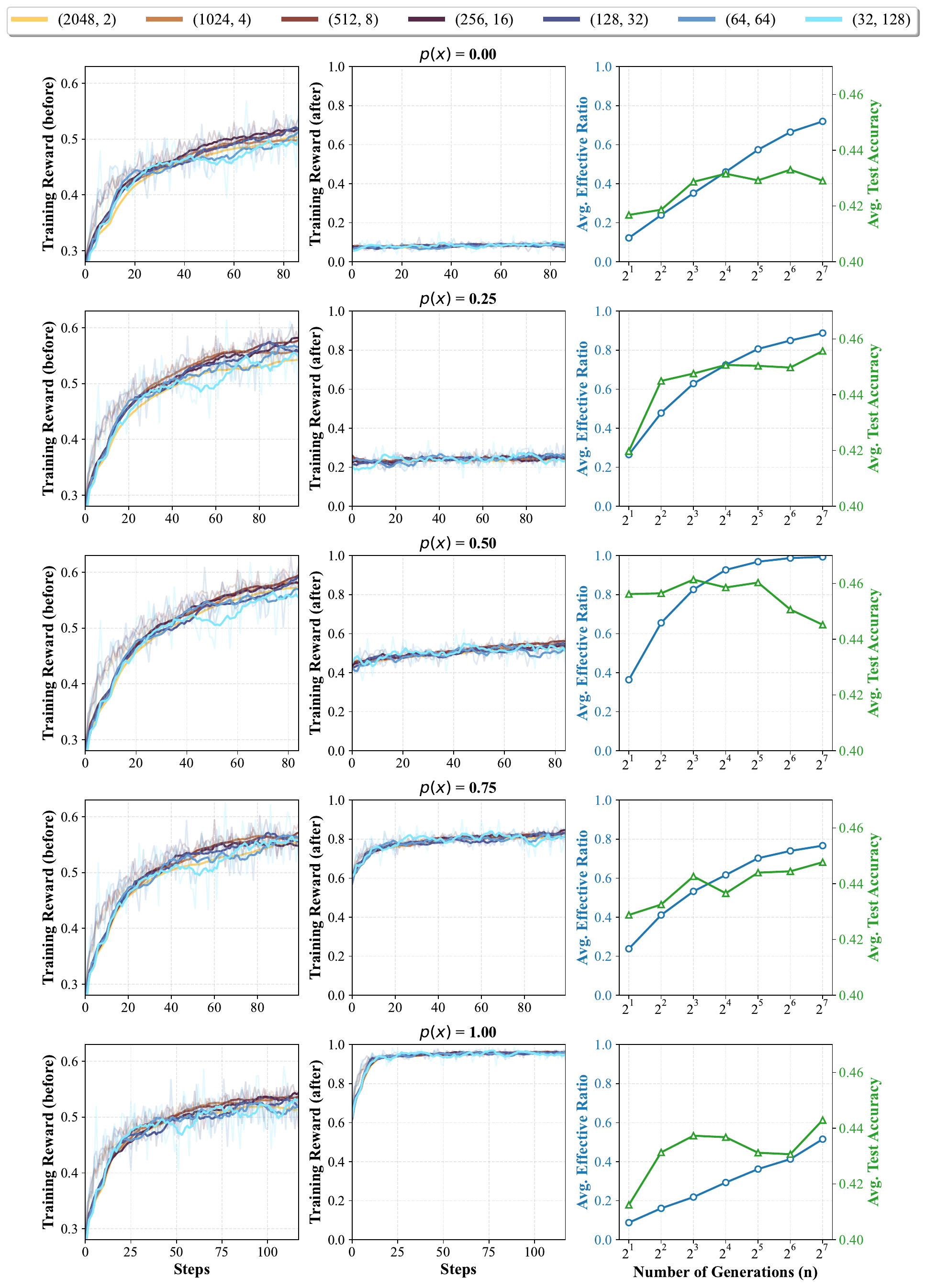}
    \vskip -0.3cm
    \caption{Results for Qwen3-4B on DeepScaleR with different $p(x)$ under different decompositions (number of prompts $m$, generations per prompt $n$), grouped by $p(x)$. (Left) Training reward before downsampling in terms of step. (Middle) Training reward after downsampling. (Right) Average effective ratio, gradient norm, and test accuracy across different thresholds.}
    \label{fig:dp_7}
\end{figure}

\begin{figure}[ht]
    \centering
    \includegraphics[trim={0 0 0 0}, clip, width=0.8\textwidth]{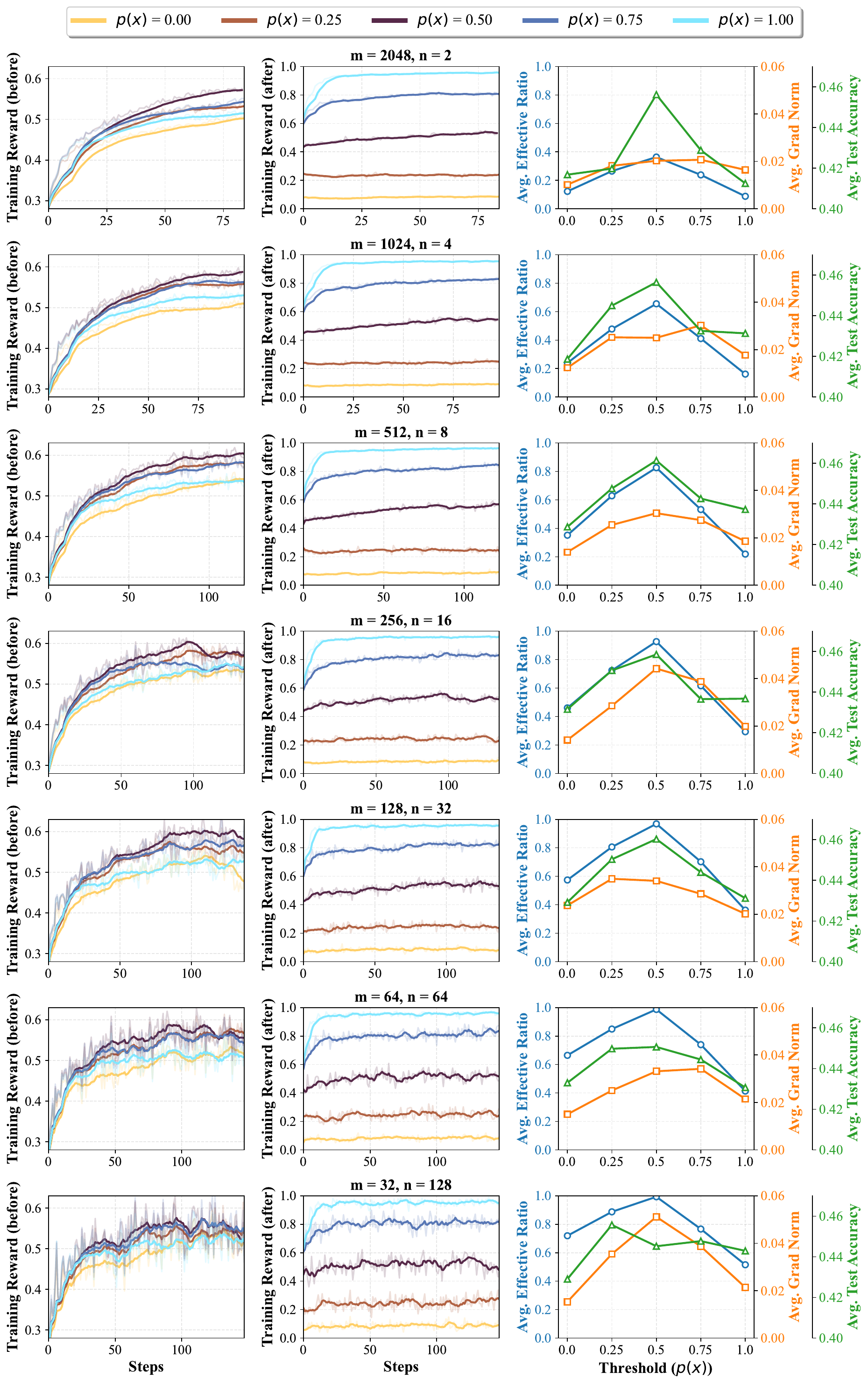}
    \vskip -0.3cm
    \caption{Results for Qwen3-4B on DeepScaleR with different $p(x)$ under different decompositions, grouped by number of prompts $m$ and generations per prompt $n$. (Left) Training reward before downsampling in terms of step. (Middle) Training reward after downsampling. (Right) Average effective ratio, gradient norm, and test accuracy across different thresholds.}
    \label{fig:dp_6}
\end{figure}

\clearpage

\begin{table}[t]\centering
\caption{{Detailed DeepScaleR Results for Fig.~\ref{fig:dp_7} and \ref{fig:dp_6}.}}
\begin{tabular}[t]{c|cc|cccccc|c} 
\midrule[0.15ex]
\multirow{2}{*}{$p(x)$} & \multirow{2}{*}{$m$} & \multirow{2}{*}{$n$} & \multirow{2}{*}{MATH500} & \multirow{2}{*}{Olymp.} & Minerva & AMC23 & AIME24 & AIME25 & \multirow{2}{*}{Avg.} \\
 & & & & & Avg@4 & Avg@32 & Avg@32 & Avg@32 & \\
\midrule[0.05ex]
/ & $\pi_\mathrm{ref}$ & \ & 65.8 & 34.4 & 26.9 & 47.3 & 10.9 & 7.1 & 32.1\\
\midrule[0.05ex]
\multirow{7}{*}{0}
& 2048 & 2 & 82.8 & 44.8 & 39.4 & 54.3 & 15.3 & 13.4 & 41.7 \\
& 1024 & 4 & 81.8 & 44.7 & 38.4 & 54.8 & 17.1 & 14.5 & 41.9 \\
& 512 & 8 & 82.6 & 48.1 & 40.0 & 57.5 & 14.7 & 14.4 & 42.9\\
& 256 & 16 & 83.4 & 46.6 & 39.4 & 58.0 & 16.7 & 14.9 & 43.2\\
& 128 & 32 & 79.8 & 45.1 & 38.1 & 59.4 & 18.2 & 16.9 & 42.9\\
& 64 & 64 & 83.4 & 47.3 & 40.7 & 59.3 & 15.4 & 13.6 & 43.3\\
& 32 & 128 & 81.0 & 49.1 & 38.5 & 54.8 & 17.7 & 16.2 & 42.9\\
\midrule[0.05ex]
\multirow{7}{*}{0.25}
& 2048 & 2 & 80.6 & 46.9 & 39.4 & 58.3 & 15.5 & 11.1 & 42.0\\
& 1024 & 4 & 84.2 & 48.2 & 39.9 & 61.6 & 16.6 & 16.5 & 44.5 \\
& 512 & 8 & 81.8 & 49.3 & 40.1 & 61.2 & 20.7 & 15.5 & 44.8 \\
& 256 & 16 & 82.8 & 47.0 & 38.5 & 63.2 & 19.6 & 19.3 & 45.1 \\
& 128 & 32 & 81.0 & 50.1 & 40.1 & 63.0 & 21.6 & 14.4 & 45.0 \\
& 64 & 64 & 83.4 & 49.1 & 41.6 & 58.8 & 19.1 & 17.8 & 45.0 \\
& 32 & 128 & 84.0 & 49.1 & 40.0 & 61.9 & 19.5 & 19.0 & 45.6 \\
\midrule[0.05ex]
\multirow{7}{*}{0.5}
& 2048 & 2 & 83.2 & 50.7 & 40.0 & 60.9 & 19.2 & 19.7 & 45.6 \\
& 1024 & 4 & 81.4 & 53.7 & 40.5 & 63.7 & 17.2 & 17.3 & 45.6 \\
& 512 & 8 & 84.2 & 50.1 & 40.3 & 64.6 & 21.5 & 16.1 & 46.1 \\
& 256 & 16 & 82.8 & 49.7 & 41.5 & 62.4 & 19.5 & 19.3 & 45.9 \\
& 128 & 32 & 84.4 & 51.8 & 39.8 & 64.4 & 20.1 & 15.7 & 46.0 \\
& 64 & 64 & 83.6 & 48.8 & 39.4 & 61.9 & 19.9 & 16.8 & 45.1 \\
& 32 & 128 & 81.6 & 45.7 & 40.8 & 62.0 & 20.1 & 16.9 & 44.5 \\
\midrule[0.05ex]
\multirow{7}{*}{0.75}
& 2048 & 2 & 81.0 & 48.7 & 39.0 & 56.6 & 17.5 & 14.6 & 42.9 \\
& 1024 & 4 & 82.0 & 49.7 & 38.6 & 58.3 & 16.5 & 14.5 & 43.3 \\
& 512 & 8 & 81.6 & 47.6 & 40.6 & 60.9 & 16.6 & 18.2 & 44.3 \\
& 256 & 16 & 81.4 & 46.1 & 39.1 & 64.0 & 17.2 & 14.2 & 43.7 \\
& 128 & 32 & 81.6 & 50.0 & 40.3 & 63.6 & 15.0 & 15.9 & 44.4 \\
& 64 & 64 & 81.0 & 50.0 & 41.1 & 61.9 & 15.1 & 17.6 & 44.4 \\
& 32 & 128 & 82.0 & 49.3 & 40.6 & 62.0 & 17.7 & 17.1 & 44.8 \\
\midrule[0.05ex]
\multirow{7}{*}{1}
& 2048 & 2 & 78.6 & 46.6 & 41.6 & 55.8 & 14.3 & 10.6 & 41.3 \\
& 1024 & 4 & 80.8 & 48.8 & 38.3 & 60.2 & 16.7 & 14.0 & 43.1 \\
& 512 & 8 & 82.4 & 45.7 & 39.9 & 59.0 & 18.8 & 16.7 & 43.7 \\
& 256 & 16 & 82.0 & 46.7 & 40.6 & 58.8 & 18.4 & 15.5 & 43.7 \\
& 128 & 32 & 81.4 & 48.2 & 39.3 & 57.6 & 17.0 & 15.2 & 43.1 \\
& 64 & 64 & 81.2 & 46.9 & 40.7 & 56.8 & 18.8 & 14.1 & 43.1 \\
& 32 & 128 & 82.4 & 45.5 & 39.7 & 62.9 & 16.6 & 18.6 & 44.3 \\
\midrule[0.15ex]
\end{tabular}
\end{table}

\clearpage
\section{\texorpdfstring{Connection between $p(x)$ and Gradient Magnitude}{Connection between p theta(x) and Gradient Magnitude}}
\label{app:p_x_grad_norm}

We now analyze the squared norm of the gradient in \cref{eq:pg_adv_x}:
\begin{align}
    ||\nabla_\theta J_x(\theta)||_2= ||\mathbb{E}_{y\sim \pi_\theta(\cdot | x)} [A(x, y)\nabla_\theta \log \pi_\theta (y|x)]||_2.
\end{align}
Using Jensen inequality, we have:
\begin{align}
    ||\mathbb{E}_{y\sim \pi_\theta(\cdot | x)} [A(x, y)\nabla_\theta \log \pi_\theta (y|x)]||_2 &\leq \mathbb{E}_{y\sim \pi_\theta(\cdot | x)} [||A(x, y)\nabla_\theta \log \pi_\theta (y|x)||_2] \\
    &= \mathbb{E}_{y\sim \pi_\theta(\cdot | x)} [|A(x, y)| \ ||\nabla_\theta \log \pi_\theta (y|x)||_2].
\end{align}
Apply Cauchy-Schwarz:
\begin{align}
    \mathbb{E}_{y\sim \pi_\theta(\cdot | x)} [|A(x, y)| \ ||\nabla_\theta \log \pi_\theta (y|x)||_2] \leq \sqrt{\mathbb{E}_{y\sim \pi_\theta(\cdot | x)} [A(x, y)^2]}\sqrt{\mathbb{E}_{y\sim \pi_\theta(\cdot | x)} [||\nabla_\theta \log \pi_\theta (y|x)||_2^2]}.
\end{align}
Now let's derive for $\mathbb{E}_{y\sim \pi_\theta(\cdot | x)} [A(x, y)^2]$:
\begin{align}
    \mathbb{E}_{y\sim \pi_\theta(\cdot | x)} [A(x, y)^2] &= \mathbb{E}_{y\sim \pi_\theta(\cdot | x)} [(r(x, y) - p_{\pi_\theta}(x))^2] \\
    &= p_{\pi_\theta}(x)(1-p_{\pi_\theta}(x))^2+(1-p_{\pi_\theta}(x))(0-p_{\pi_\theta}(x))^2 \\
    &= p_{\pi_\theta}(x)(1-p_{\pi_\theta}(x))
\end{align}
which is maximized at $p_{\pi_\theta}(x)=\frac{1}{2}$. Therefore, intermediate-level prompts yield the largest expected magnitude of advantage and upper bound on gradient updates. These observations motivate a prompt curriculum strategy that prioritizes prompts with intermediate difficulty ($p_{\pi_\theta}(x)=\frac{1}{2}$) to enhance training efficiency.

\clearpage
\section{Experiment Details}
\label{app:main_exp_detail}

\subsection{Baselines Algorithms}

We list the pseudo-code for GRPO, Pre-filter, DS, and SPEED. For the pseudo-code of GRESO, please refer to \citet{zheng2025actpaysefficientreinforcement}.

\begin{algorithm}
\caption{GRPO}
\label{alg:vanilla}
\begin{algorithmic}[1]
\Require Number of prompts $m$, generations per prompt $n$
\State Initialize policy $\pi_0$
\For{$t=0$ to $T-1$}
  \State Sample a batch with $m$ prompts: $\mathcal{D}_m=\{x^i\}_{i=1}^m \subset \mathcal{D}$.
  \State Generate $n$ responses: $\mathcal{D}_m = \bigl\{(x^i,\{y^{i,j}\}_{j=1}^n) \bigr\}_{i=1}^m$ where $y^{i,j} \overset{\mathrm{iid}}{\sim} \pi_t(\cdot\mid x^i)$.
  \State Update to $\pi_{t+1}$ with GRPO objective using $\mathcal{D}_m$.
\EndFor
\end{algorithmic}
\end{algorithm}

\begin{algorithm}
\caption{Pre-filter}
\label{alg:pf}
\begin{algorithmic}[1]
\Require Number of prompts $m$, generations per prompt $n$, pre-filter generations $n_\mathrm{pre}$, thresholds $p_\mathrm{low}$ and $p_\mathrm{high}$
\State Initialize policy $\pi_0$
\State Generate $n_\mathrm{pre}$ responses for each prompt in the dataset: $\{y^j\}_{j=1}^{n_\mathrm{pre}}\sim\pi_0(\cdot|x)$ for each $x \in \mathcal{D}$
\State Filter to keep prompts with accuracy between $p_\mathrm{low}$ and $p_\mathrm{high}$: 
$$\mathcal{D} \gets \{x\in \mathcal{D} \mid p_{low} < \frac{1}{n_\mathrm{pre}}\sum_{j=1}^{n_\mathrm{pre}} r(x, y^j) < p_{high}\}$$
\For{$t=0$ to $T-1$}
  \State Sample a batch with $m$ prompts: $\mathcal{D}_m=\{x^i\}_{i=1}^m \subset \mathcal{D}$.
  \State Generate $n$ responses: $\mathcal{D}_m = \bigl\{(x^i,\{y^{i,j}\}_{j=1}^n) \bigr\}_{i=1}^m$ where $y^{i,j} \overset{\mathrm{iid}}{\sim} \pi_t(\cdot\mid x^i)$.
  \State Update to $\pi_{t+1}$ with GRPO objective using $\mathcal{D}_m$.
\EndFor
\end{algorithmic}
\end{algorithm}

\begin{algorithm}
\caption{Dynamic-sampling (DS)}
\label{alg:ds}
\begin{algorithmic}[1]
\Require Number of prompts $m$, generations per prompt $n$, sampling parameter $k$
\State Initialize policy $\pi_0$, $\mathcal{D}_\text{buffer}\gets\emptyset$
\For{$t=0$ to $T-1$}
    \While{$|\mathcal{D}_\text{buffer}| < m$}
        \State Sample a batch with $km$ prompts: $\mathcal{D}_{km}=\{x^i\}_{i=1}^{km} \subset \mathcal{D}$.
        \State Generate $n$ responses: $\mathcal{D}_{km} = \bigl\{(x^i,\{y^{i,j}\}_{j=1}^n) \bigr\}_{i=1}^{km}$ where $y^{i,j} \overset{\mathrm{iid}}{\sim} \pi_t(\cdot\mid x^i)$
        \State Select prompts with mean reward between 0 and 1:
        $$\mathcal{D}_\text{buffer}\gets \mathcal{D}_\text{buffer}\cup \{(x,\{y^j\}_{j=1}^n)\in \mathcal{D}_{km} \mid 0 < \frac{1}{n}\sum_{j=1}^n r(x, y^j) < 1\}$$
    \EndWhile
        \State Sample a batch with $m$ prompts: $\mathcal{D}_m=\bigl\{(x^i,\{y^{i, j}\}_{j=1}^n) \bigr\}_{i=1}^m \subset \mathcal{D}_\text{buffer}$.
        \State $\mathcal{D}_\text{buffer}\gets \emptyset$
        \State Update to $\pi_{t+1}$ with GRPO objective using $\mathcal{D}_m$.
\EndFor
\end{algorithmic}
\end{algorithm}

\begin{algorithm}
\caption{SPEED}
\label{alg:speed}
\begin{algorithmic}[1]
\Require Number of prompts $m$, generations per prompt $n$, sampling parameter $k$, screening number of responses $n_{\mathrm{init}}$ ($n \geq n_{\mathrm{init}}$)
\State Initialize policy $\pi_0$, $\mathcal{D}_\text{buffer}\gets\emptyset$, $\mathcal{D}_\text{accepted}\gets\emptyset$
\For{$t=0$ to $T-1$}
    \While{$|\mathcal{D}_\text{buffer}| < m$}
        \State Sample a batch with $km$ prompts: $\mathcal{D}_{km}=\{x^i\}_{i=1}^{km} \subset \mathcal{D}$.
        \State Generate $n_{\mathrm{init}}$ times for $\mathcal{D}_{km}$ and $n-n_{\mathrm{init}}$ times for $\mathcal{D}_\text{accepted}$: 
        \Statex \hspace{\algorithmicindent} \hspace{\algorithmicindent} \hspace{\algorithmicindent} $\mathcal{D}_{km} = \bigl\{(x^i,\{y^{i,j}\}_{j=1}^{n_{\mathrm{init}}}) \bigr\}_{i=1}^{km}$ where $y^{i,j} \overset{\mathrm{iid}}{\sim} \pi_t(\cdot\mid x^i)$
        \Statex \hspace{\algorithmicindent} \hspace{\algorithmicindent} \hspace{\algorithmicindent} $\mathcal{D}_\text{accepted} \gets \mathcal{D}_\text{accepted} \cup \bigl\{(x^i,\{y^{i,j}\}_{j=1}^{n-n_{\mathrm{init}}}) \bigr\}_{i=1}^{|\mathcal{D}_\text{accepted}|}$ where $y^{i,j} \overset{\mathrm{iid}}{\sim} \pi_t(\cdot\mid x^i)$.
        \State Add $\mathcal{D}_\text{accepted}$ to $\mathcal{D}_\text{buffer}$: $\mathcal{D}_\text{buffer}\gets \mathcal{D}_\text{buffer}\cup\mathcal{D}_\text{accepted}$
        \State Select prompts with mean reward between 0 and 1 and add to $\mathcal{D}_\text{accepted}$: $$\mathcal{D}_\text{accepted} \gets \{(x,\{y^j\}_{j=1}^{n_{\mathrm{init}}})\in \mathcal{D}_{km} \mid 0 < \frac{1}{n_{\mathrm{init}}}\sum_{j=1}^{n_{\mathrm{init}}} r(x, y^j) < 1\}$$
    \EndWhile
        \State Sample a batch with $m$ prompts: $\mathcal{D}_m=\bigl\{(x^i,\{y^{i,j}\}_{j=1}^n) \bigr\}_{i=1}^m \subset \mathcal{D}_\text{buffer}$.
        \State $\mathcal{D}_\text{buffer}\gets \mathcal{D}_\text{buffer}\setminus \mathcal{D}_m$
        \State Update to $\pi_{t+1}$ with GRPO objective using $\mathcal{D}_m$.
\EndFor
\end{algorithmic}
\end{algorithm}

\clearpage
\subsection{Dataset, Model, Reward, Evaluation Details}

We adopt the exact same setting in our preliminary investigation. Refer to Appendix~\ref{app:exp_detail} for details on datasets, models, rewards, and evaluations.

\subsection{Hyperparameters}

We list the hyperparameters for each method below. We tune the sample batch size / sampling parameter for DS, SPEED, GRESO, and \alg{} with the \textbf{bolded} one as the best one. For \alg{}, we always use the same-sized model as the value model for the main results. The learning rate of the value model in \alg{} is tuned within $\{1e-6, 3e-6, 1e-5\}$. All learning rates make the value model converge to the same performance but larger learning rates are more unstable than the smaller one. Therefore, we pick 1e-6 as the learning rate for the value model. We also tune the sampling parameter $k$ from $\{2, 4, 8\}$ and observe that it has a minor effect on the accuracy of the value model. Larger values of $k$ lead to higher effective ratios, as the filtering becomes more aggressive. However, increasing $k$ beyond 4 yields only marginal improvements in the effective ratio. Therefore, we set $k = 4$ for all experiments on \alg. An ablation on the value model size is included in Appendix~\ref{app:value_model_size}.

\begin{table*}[htb!]\centering
\resizebox{0.9\linewidth}{!}{
\begin{tabular}{p{0.2\linewidth}p{0.2\linewidth}p{0.3\linewidth}p{0.3\linewidth}}
\midrule[0.3ex]
\textbf{Dataset} & \textbf{Method} & 
\textbf{Parameters} & \\
\midrule[0.15ex]
MATH & 
GRPO &
$m=512$ & $n=16$ \\
& & $lr=8e-6$ \\
\midrule[0.15ex]
DeepScaleR & 
GRPO &
$m=512$ & $n=16$ \\
& & $lr=4e-6$ \\
\midrule[0.15ex]
MATH & 
Pre-filter &
$m=512$ & $n=16$ \\
& & $lr=8e-6$ & $n_\mathrm{pre}=16$ \\
& & $p_\mathrm{low}=0$ & $p_\mathrm{high}=1$ \\
\midrule[0.15ex]
DeepScaleR & 
Pre-filter &
$m=512$ & $n=16$ \\
& & $lr=4e-6$ & $n_\mathrm{pre}=16$ \\
& & $p_\mathrm{low}=0$ & $p_\mathrm{high}=1$ \\
\midrule[0.15ex]
MATH & 
DS &
$m=512$ & $n=16$ \\
& & $lr=8e-6$ & $k=1/\textbf{2}/4$ \\
\midrule[0.15ex]
DeepScaleR & 
DS &
$m=512$ & $n=16$ \\
& & $lr=4e-6$ & $k=1/\textbf{2}/4$ \\
\midrule[0.15ex]
MATH & 
SPEED &
$m=512$ & $n=16$ \\
& & $lr=8e-6$ & $k=1/\textbf{2}/4$ \\
& & $n_\mathrm{init}=\textbf{4}/8$ \\
\midrule[0.15ex]
DeepScaleR & 
SPEED &
$m=512$ & $n=16$ \\
& & $lr=4e-6$ & $k=1/\textbf{2}/4$ \\
& & $n_\mathrm{init}=\textbf{4}/8$ \\
\midrule[0.15ex]
MATH & GRESO & $m=512$ & $n=16$ \\
& & $lr=8e-6$ & $B_\mathrm{r}^\mathrm{default}=\textbf{768}/1024$ \\
& & $p_{easy}=0.5$ & $p_{hard}=0.5$ \\
& & $\alpha_{easy} = 0.083$ & $\alpha_{hard} = 0.167$ \\
& & $\Delta p=0.01$ \\
\midrule[0.15ex]
DeepScaleR & GRESO & $m=512$ & $n=16$ \\
& & $lr=4e-6$ & $B_\mathrm{r}^\mathrm{default}=\textbf{768}/1024$ \\
& & $p_{easy}=0.5$ & $p_{hard}=0.5$ \\
& & $\alpha_{easy} = 0.083$ & $\alpha_{hard} = 0.167$ \\
& & $\Delta p=0.01$ \\
\midrule[0.15ex]
MATH & \alg & $m=512$ & $n=16$ \\
& & $lr=8e-6$ & $lr_\mathrm{critic}$ = \textbf{1e-6}/3e-6/1e-5 \\
& & $\tau=0.5$ & $k=2/\textbf{4}/8$ \\
\midrule[0.15ex]
DeepScaleR & \alg & $m=512$ & $n=16$ \\
& & $lr=4e-6$ & $lr_\mathrm{critic}$ = \textbf{1e-6}/3e-6/1e-5 \\
& & $\tau=0.5$ & $k=2/\textbf{4}/8$ \\
\midrule[0.3ex]
\end{tabular}}
\end{table*}

\clearpage
\section{Value Model Size Ablation}
\label{app:value_model_size}

\begin{figure}[h]
    \centering
    \includegraphics[trim={0 0 0 0}, clip, width=\textwidth]{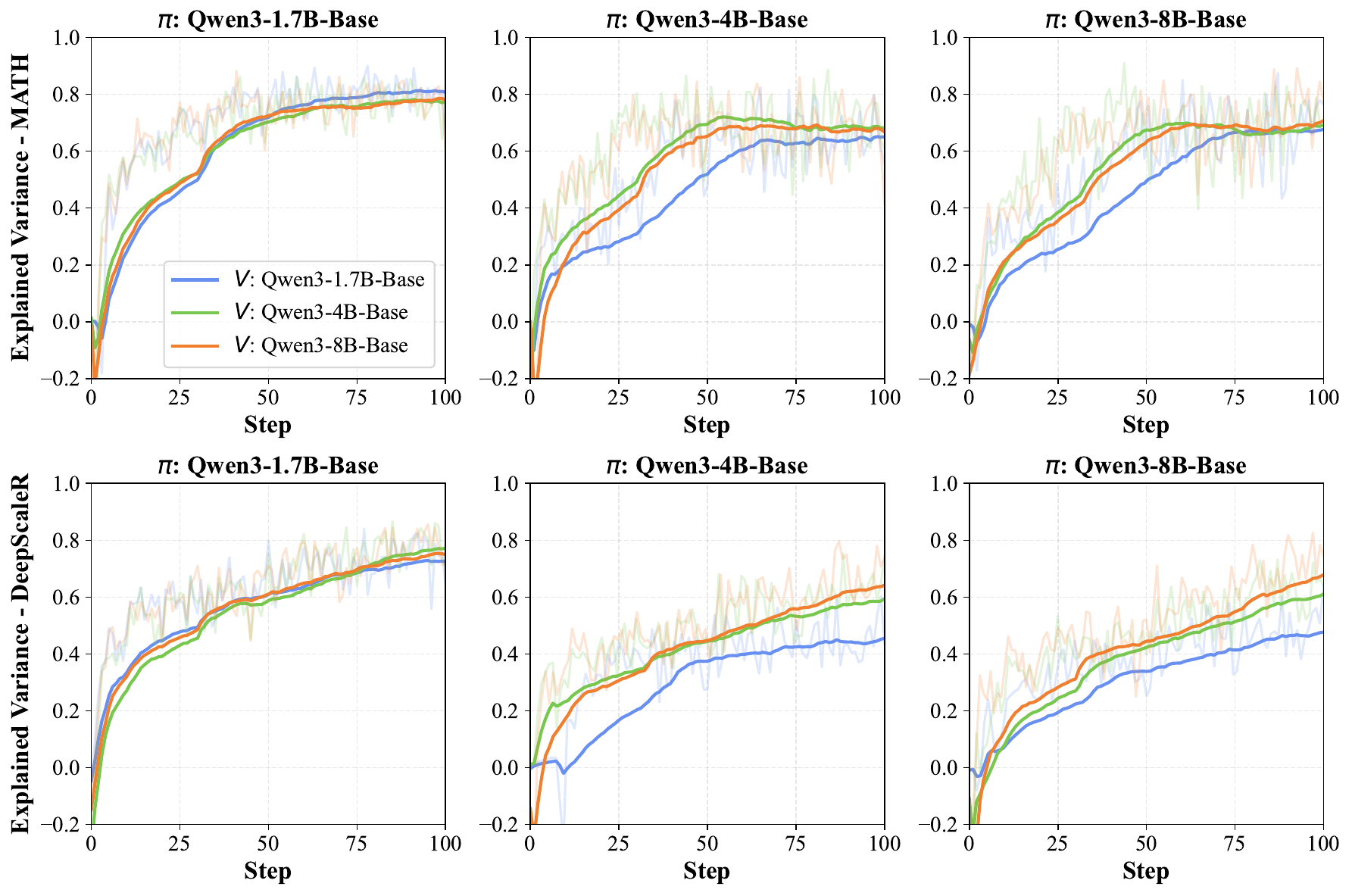}
    \vskip -0.4cm
    \caption{Explained variance on MATH and DeepScaleR with different combinations of Qwen3 base models (1.7B / 4B / 8B) for policy ($\pi$) and the value model ($V$).}
    \label{fig:explained_variance}
\end{figure}

We ablate the value model used in \alg{} across three different model sizes, with results on explained variance of the value prediction shown in Figure~\ref{fig:explained_variance}. Similar to Section~\ref{sec:experiment}, we use the average reward of 16 generations for each prompt as the ground-truth $p(x)$. Overall, larger value models exhibit faster convergence compared to smaller ones. On the MATH dataset, all three value models eventually converge to similar performance levels. However, on the larger DeepScaleR dataset, we observe a substantial gap after 100 training steps: the smaller value model significantly underperforms relative to its larger counterparts.

We hypothesize that the smaller value model may require more training steps to reach comparable accuracy and that, given sufficient time, all models could eventually converge to a similar point. Nonetheless, this result highlights the benefit of larger value models in the early stages of training, especially on large-scale datasets.

\end{document}